\documentclass[11pt]{article}

\usepackage[final]{acl}

\usepackage{times}
\usepackage{latexsym}

\usepackage[T1]{fontenc}
\usepackage[utf8]{inputenc}

\usepackage{microtype}

\usepackage{inconsolata}

\usepackage{graphicx}

\usepackage{stfloats}

\usepackage{hyperref}       % hyperlinks
\usepackage{url}            % simple URL typesetting
\usepackage{booktabs}       % professional-quality tables
\usepackage{amsfonts}       % blackboard math symbols
\usepackage{nicefrac}       % compact symbols for 1/2, etc.
\usepackage{microtype}      % microtypography
\usepackage{xcolor}         % colors

\usepackage{multirow}
\usepackage{inconsolata}
\usepackage{caption}
\usepackage{graphicx} 
\usepackage[most]{tcolorbox}
\usepackage{amsmath}
\usepackage{titletoc}

\definecolor{mycolor_blue}{HTML}{E7EFFA}
\definecolor{mycolor_green}{HTML}{E6F8E0}
\definecolor{mycolor_gray}{HTML}{ECECEC}
\definecolor{mycolor_red}{HTML}{FFE6E6}
\definecolor{mycolor_yellow}{HTML}{FFFFCC}
\definecolor{mycolor_purple}{HTML}{E6E6FF}
\definecolor{softgray}{RGB}{242, 242, 242}  % 柔和的浅灰色
\definecolor{softblue}{RGB}{208, 231, 255}  % 柔和的蓝色
\definecolor{skyblue}{RGB}{135, 206, 235}
\definecolor{lightblue}{RGB}{173, 216, 230}
\definecolor{lightyellow}{RGB}{255, 255, 224}

\definecolor{BlueGreen}{RGB}{6, 180, 185} 
\definecolor{RedOrange}{RGB}{240, 98, 51}
\newcommand{\blue}[1]{$_{\color{BlueGreen}\downarrow #1}$}
\newcommand{\red}[1]{$_{\color{RedOrange}\uparrow #1}$}

\newcommand{\homepage}{\raisebox{-1.5pt}{\includegraphics[height=1.05em]{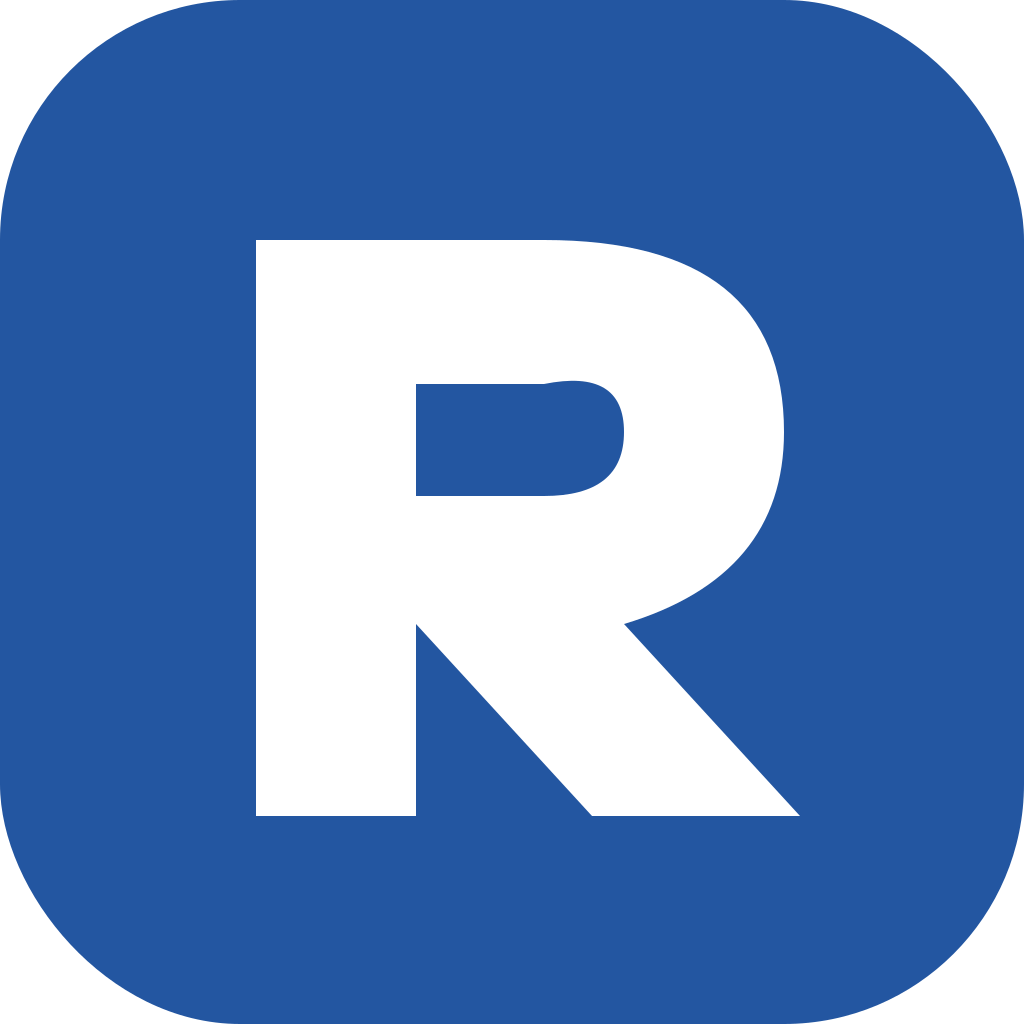}}\hspace{0.3em}}
\newcommand{\github}{\raisebox{-1.5pt}{\includegraphics[height=1.05em]{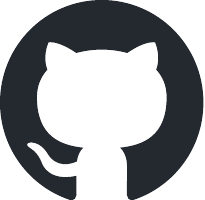}}\hspace{0.3em}}
\newcommand{\benchmark}{\raisebox{-1.5pt}{\includegraphics[height=1.05em]{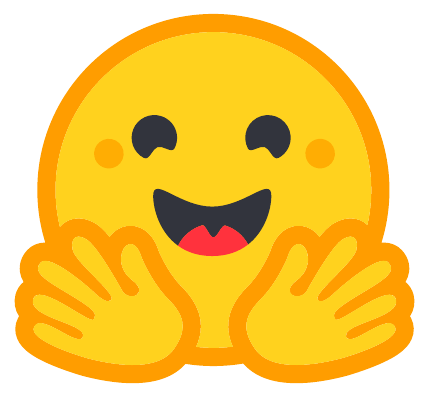}}\hspace{0.3em}}
\newcommand{\leaderboard}{\raisebox{-1.5pt}{\includegraphics[height=1.05em]{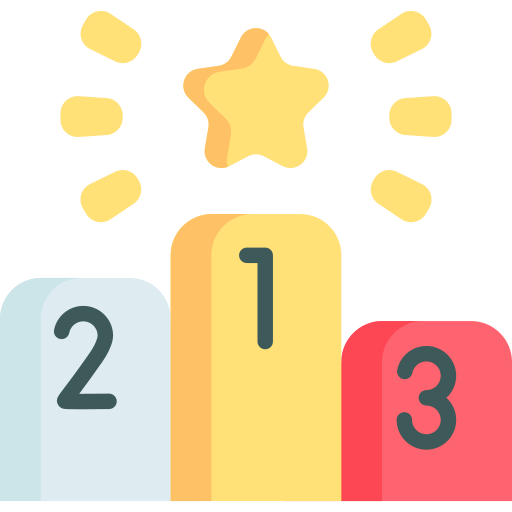}}\hspace{0.3em}}
\newcommand{\dataset}{\raisebox{-1.5pt}{\includegraphics[height=1.05em]{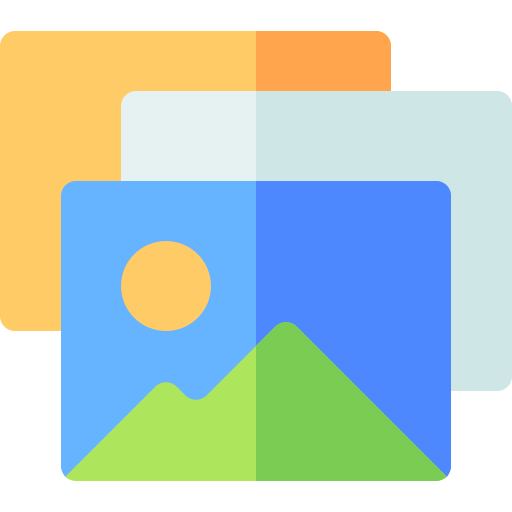}}\hspace{0.3em}}

\title{T2I-ReasonBench: Benchmarking Reasoning-Informed \\
Text-to-Image Generation}

\author{
  \bf{Kaiyue Sun}\textsuperscript{1} \quad 
  \bf{Rongyao Fang}\textsuperscript{2} \quad 
  \bf{Chengqi Duan}\textsuperscript{1}  \quad
  \bf{Xian Liu}\textsuperscript{2} \quad  
  \bf{Aoxue Li}\textsuperscript{3}  \quad 
  \bf{Xihui Liu}\textsuperscript{1$\dag$} \\
  \\
\textsuperscript{1}HKU  \quad
\textsuperscript{2}CUHK  \quad 
\textsuperscript{3}Huawei Noah’s Ark Lab \\
\\
{\homepage \href{https://kaiyuesun98.github.io/T2I-ReasonBench-website/}{{\text{Homepage}}}} \quad 
{\leaderboard \href{https://kaiyuesun98.github.io/T2I-ReasonBench-website/\#leaderboard}{{\text{Leaderboard}}}} \quad 
{\github \href{https://github.com/KaiyueSun98/T2I-ReasonBench}{{\text{Code}}}}
\quad 
{\benchmark \href{https://huggingface.co/datasets/Kaiyue/T2I-ReasonBench}{{\text{Benchmark}}}}
\quad 
{\dataset \href{https://huggingface.co/datasets/Kaiyue/T2I-ReasonBench-Images}{{\text{Dataset}}}}
}

\begin{document}

\twocolumn[{%
\renewcommand\twocolumn[1][]{#1}%
\maketitle
\vspace{-2em}

\includegraphics[width=\linewidth]{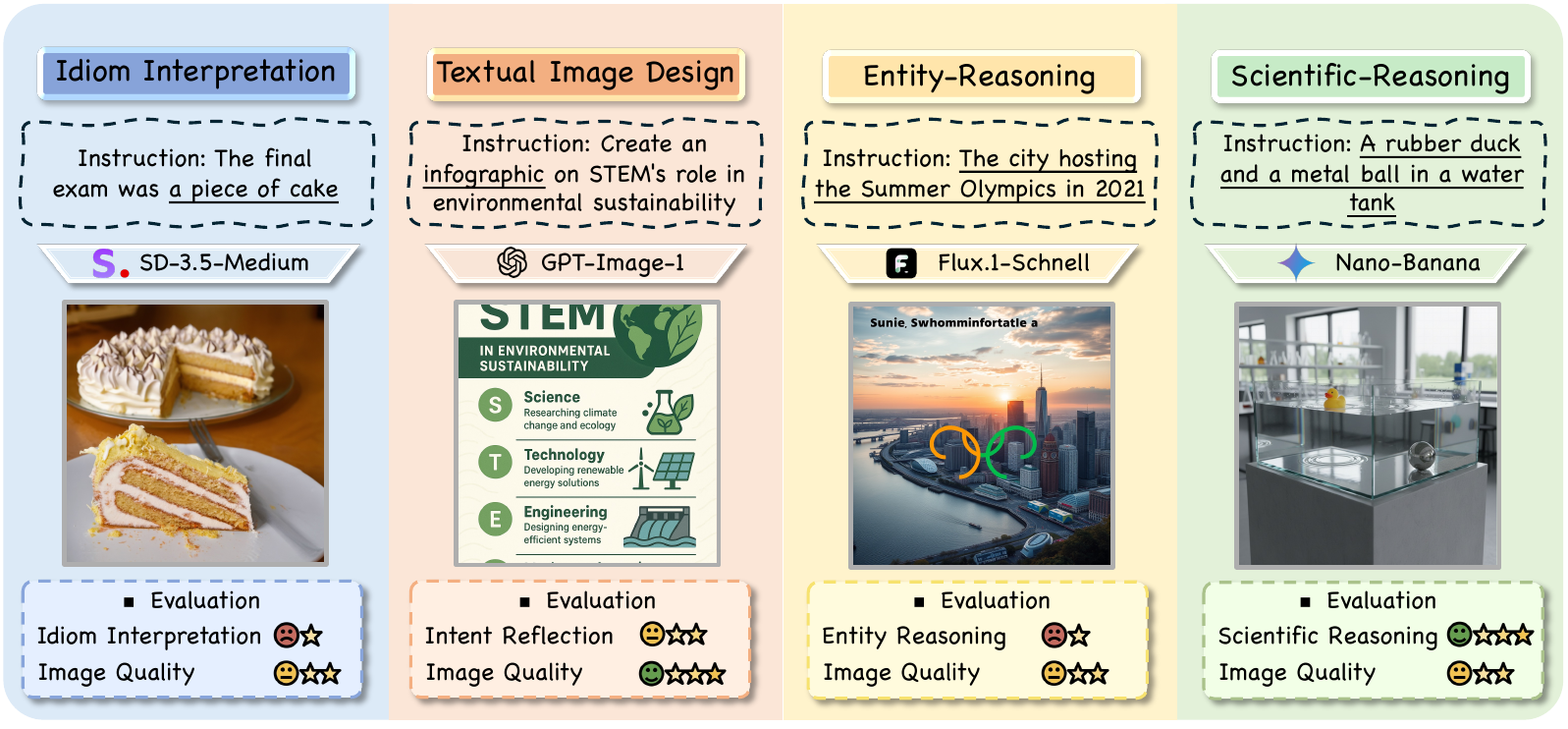}
\captionof{figure}{\textbf{Overview of T2I-ReasonBench.} We propose T2I-ReasonBench, a benchmark evaluating reasoning-informed generation of text-to-image (T2I) models. It consists of four dimensions: \textbf{Idiom Interpretation, Textual Image Design, Entity-Reasoning and Scientific-Reasoning}. We propose a two-stage evaluation protocol to measure T2I-ReasonScore, a metric that integrates reasoning accuracy, detail faithfulness and image quality. We benchmark various types T2I models, and provide comprehensive analysis on their reasoning and generation abilities. 
}

\label{fig:teaser}
}]

% \maketitle

\renewcommand{\thefootnote}{\fnsymbol{footnote}}
\footnotetext[2]{Corresponding Author}

\begin{abstract}
Text-to-image (T2I) generative models have achieved remarkable progress, demonstrating exceptional capability in synthesizing high-quality images from textual prompts. While existing research and benchmarks have extensively evaluated the ability of T2I models to follow the literal meaning of prompts, their ability to reason over prompts with domain knowledge to uncover implicit meaning and contextual nuances remains underexplored. To bridge this gap, we introduce T2I-ReasonBench, a novel benchmark designed to explore the knowledge-driven reasoning capabilities of T2I models.
T2I-ReasonBench comprises 900 meticulously designed prompts organized into four dimensions: \textbf{(1) Idiom Interpretation}, \textbf{(2) Textual Image Design}, \textbf{(3) Entity Reasoning}, and \textbf{(4) Scientific Reasoning}. These dimensions challenge models to integrate domain knowledge, infer implicit meaning, and resolve contextual ambiguities. To quantify the performance, we introduce a two-stage evaluation framework: a large language model (LLM) generates prompt-specific question-criterion pairs that evaluate if the image includes the essential elements resulting from correct reasoning; a multimodal LLM (MLLM) then scores the generated image against these criteria. 
Our comprehensive study across 16 state-of-the-art diffusion and unified multimodal models (UMMs) reveals two primary bottlenecks. First, many models lack the foundational reasoning ability to fully comprehend complex prompts. Second, even models with stronger reasoning modules exhibit a persistent gap between their internal understanding and the final generated image. This highlights an urgent need for the next generation of T2I systems to not only improve their reasoning capability but also to enhance integration between reasoning and synthesis. Project page: \url{https://kaiyuesun98.github.io/T2I-ReasonBench-website/}.
\end{abstract}

\section{Introduction}
\label{sec:intro}
Recent advancements in T2I generative models have enabled the creation of visually appealing images from textual prompts. However, these models often struggle with generating complex scenes that demand reasoning. 
Current benchmarks~\citep{yu2022scaling, hu2023tifa, huang2023t2icompbench, ghosh2023geneval, hu2024ella, li2024genai, wu2024conceptmix, huang2025t2icompbench++, wei2025tiif} primarily evaluate literal prompt-image alignment, focusing on object attributes (e.g., color, attribute, count) and relationships. 
While DPG-Bench~\citep{hu2024ella} extends evaluation to long-text comprehension, it remains confined to multi-object composition tasks. 
These frameworks fail to test models' ability to reason beyond explicit instructions. 
For instance, generating an image of ``A beach ball and a marble in a swimming pool'' requires not only object composition but also reasoning about physical laws (e.g., inferring the ball floats while the marble sinks). 
Such reasoning necessitates understanding related scientific knowledge, such as material density and buoyancy, as well as integrating the reasoning process into T2I generation. 

To address this gap, we propose \textbf{T2I-ReasonBench}, a novel benchmark designed to systematically evaluate the reasoning ability of T2I models in four dimensions: (1) Idiom Interpretation: deciphering the implicit meanings of idiomatic expressions with the context to generate appropriate images. (2) Textual Image Design: understanding the intention of design and effectively planning the integrated visual-textual contents. (3) Entity Reasoning: applying and integrating the knowledge about world entities in image generation, and (4) Scientific Reasoning: reasoning with scientific knowledge (e.g., physics, chemistry) to produce images adhering to the underlying scientific laws.
T2I-ReasonBench encompasses the above four dimensions with 900 meticulously designed prompts, all of which require deep reasoning.

To rigorously evaluate the performances of T2I models, we introduce a two-stage evaluation framework and propose \textbf{T2I-ReasonScore}, a quantitative metric for assessing the quality of reasoning-informed T2I generation. First, an LLM generates specific question-criterion pairs for each prompt. To evaluate the images, an MLLM then answers each question and assigns a score based on the paired criterion. By averaging these scores, we measure how faithfully the image reflects the implicit meaning of the prompt, capturing the effectiveness of model's reasoning. 
Our approach allows for fine-grained and interpretable evaluation of models' reasoning ability and addresses the limitation of previous benchmarks that focused solely on literal prompt following.

We evaluate 16 state-of-the-art T2I models, including 8 diffusion models, 5 UMMs, and 3 proprietary models. Based on the results, we present the first comprehensive study to decouple reasoning with generation and evaluate the gap in between. Our analysis reveals that reasoning is the primary bottleneck for current models. Furthermore, even though UMMs unify understanding and generation in a single framework, they still struggle to transfer their knowledge and reasoning effectively into the final image.

Our contributions are threefold:  
(1) We propose T2I-ReasonBench, a novel benchmark with meticulously designed tasks to explore the reasoning-informed T2I generation.
(2) Our prompt-specific evaluation framework enables fine-grained and interpretable evaluation.
(3) We evaluate a range of T2I systems, conduct a thorough analysis on their reasoning and generation abilities, and provide insights into limitations of current models and future model design.

\section{Related Work}
\label{sec:related_work}
\subsection{Text-to-image Generation.} 

\textbf{Diffusion models.}
T2I generation has seen rapid advances in recent years, primarily driven by the emergence and refinement of diffusion models \citep{dhariwal2021diffusion, ho2020denoising, nichol2021glide, saharia2022photorealistic}. By formulating image synthesis as a progressive denoising process, these models pushed the boundaries of quality and controllability of T2I generation, and established the backbone for modern T2I systems like the Stable Diffusion series \citep{sd3medium}, and the Flux series \citep{flux2024}. 
Recent models like HiDream~\citep{hidream} and Qwen-Image~\citep{wu2025qwen} further extend this paradigm, achieving fine-grained, photorealistic T2I generation, solidifying diffusion as the backbone of modern T2I systems.

\noindent \textbf{Unified multimodal models.}
To achieve better token-level alignment between text and image modalities, recent research has shifted toward LLM-based architectures.
This includes both autoregressive models, which synthesize images by directly predicting sequences of visual tokens\citep{ramesh2021zero, ding2021cogview, sun2024autoregressive, liu2024lumina}, and unified multimodal models \citep{chameleon2024mixed, showo, janus, deng2025emerging, chen2025blip3, got, duan2025got}.  These unified systems typically combine an autoregressive language model with a diffusion module to integrate understanding and generation. For instance, GoT \citep{got} uses an MLLM for semantic-spatial reasoning before diffusion-based synthesis, and Bagel \citep{deng2025emerging} unifies an LLM and a diffusion model within an architecture of Mixture of Transformer (MoT) to generate reasoning chains prior to image creation.

\subsection{Text-to-image Benchmarks and Evaluation Metrics.}
\textbf{Benchmarks.}
Current T2I benchmarks~\citep{yu2022scaling, hu2023tifa, huang2023t2icompbench, ghosh2023geneval, hu2024ella, li2024genai, wu2024conceptmix, huang2025t2icompbench++, wei2025tiif} primarily evaluate literal prompt-image alignment. For example,
GenEval~\citep{ghosh2023geneval} utilizes object detection techniques to test whether generated images correctly capture object co-occurrence, position, count, and color described in the prompts. 
Recent benchmarks have shifted focus from literal alignment to reasoning capabilities of T2I models. For instance, Commonsense-T2I~\citep{fu2024commonsense} tests everyday logic through adversarial prompt pairs; PhyBench \citep{meng2024phybench} evaluates physical common sense; WISE \citep{niu2025wise} assesses broader world knowledge; and R2I-Bench \citep{chen2025r2i} includes both composition and reasoning categories. 
Our benchmark encompasses a comprehensive set of reasoning dimensions and provides a deeper, more insightful analysis of models' reasoning and generation abilities.

\noindent \textbf{Evaluation Metrics.} Conventional text-image alignment metrics like CLIPscore~\citep{hessel2021clipscore} and VQAscore~\citep{lin2024evaluating} work as bag-of-words models, lacking the expertise needed to evaluate specific composition and reasoning generations. To address this, many works adopt a more targeted, disentangled question-answering framework. Leveraging powerful LLMs and MLLMs, this method first generates specific diagnostic questions and then answers them by inspecting the image. This approach has been successfully applied across studies evaluating text-image alignment~\citep{hu2023tifa, dsg, vq2, vpeval}, composition~\citep{wu2024conceptmix}, reasoning~\citep{chen2025r2i}, and factual correctness~\citep{hallu}. 
Recent work has also examined the reliability of MLLMs as evaluators themselves: MLLM-as-a-Judge~\citep{chen2024mllmjudge} and A-Bench~\citep{zhang2025abench} benchmark MLLM judgments against human annotations across scoring, semantic alignment, and perceptual quality, while MJ-Bench~\citep{chen2025mjbench} studies fine-grained preference judgments for T2I generation. MJ-Video~\citep{tong2025mjvideo} further extends this fine-grained judging paradigm to video generation, motivating explicit criteria and human-aligned validation for multimodal evaluation.
Given the complexity of our benchmark, this targeted, disentangled approach is more reliable than metrics with identical instructions.

\section{Benchmark Construction}
\label{sec:dataset}

\begin{figure*}

\begin{minipage}{0.31\textwidth}
    \centering        
    \includegraphics[width=\linewidth]{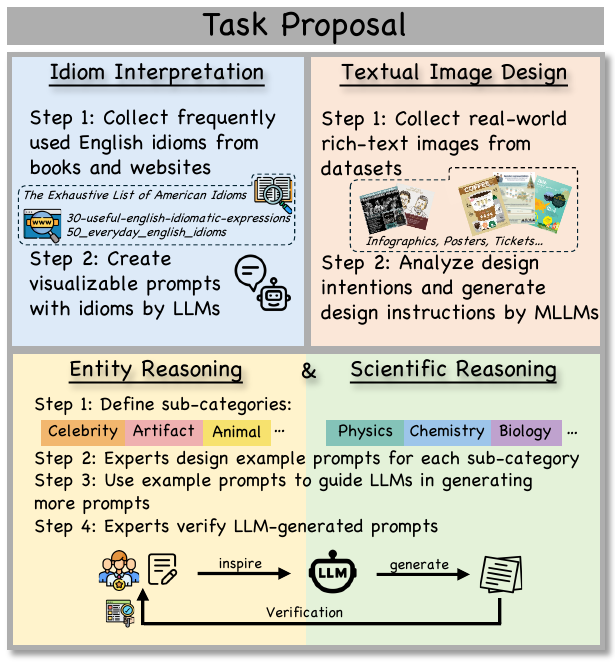}   
\end{minipage}
\begin{minipage}{0.33\textwidth}
    \centering      
    \includegraphics[width=\linewidth]{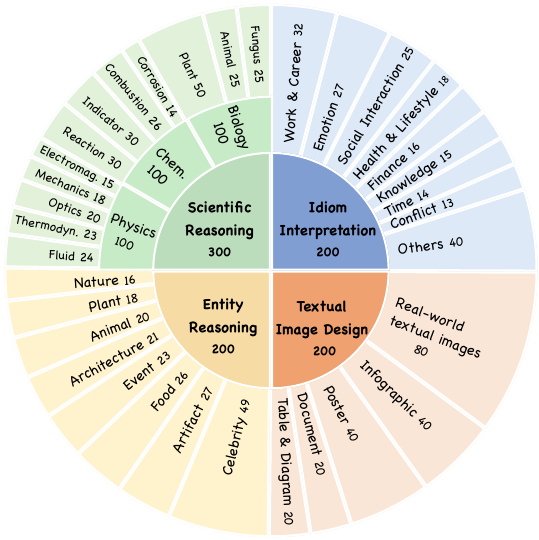} 
\end{minipage}
\begin{minipage}{0.31\textwidth}
    \centering      
    \includegraphics[width=\linewidth]{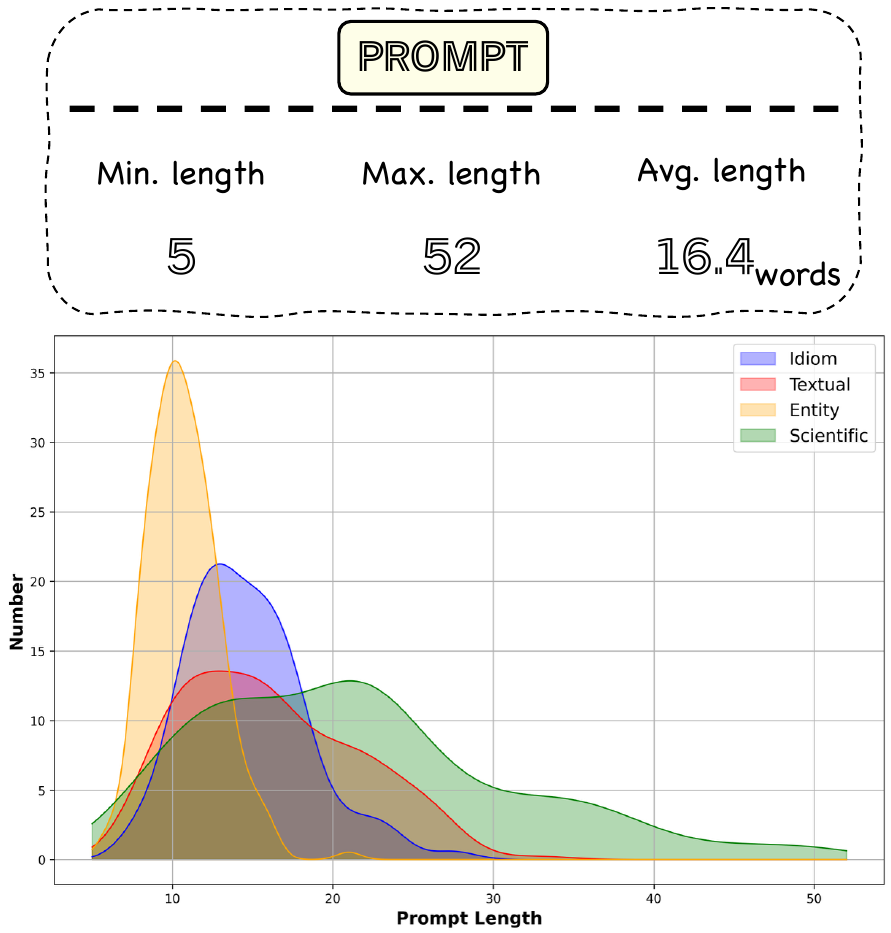} 
\end{minipage}

\caption{\textit{\textbf{Left: Prompt Generation Pipeline. }}
Idiom Interpretation: collect idioms from established sources and use LLMs to create prompts.
Textual Image Design: source real-world rich-text images, use MLLMs to analyze design intentions and generate corresponding instructions.
Entity \& Scientific Reasoning: A Human-in-the-loop process: define subcategories, expert prompt design, LLM expansion into larger sets, and final expert verification to ensure accuracy.
\textit{\textbf{Middle: Distribution of Subcategories.}}
 The inner ring represents the four dimensions, the outer rings display subcategories and their sample counts.
\textit{\textbf{Right: Prompt Suite Statistics. }}
Top: Statistics about prompt length.
Bottom: The distribution of prompt lengths across the four dimensions. } 
\label{fig:prompt_suite} 
\end{figure*}

While modern T2I models are good at explicit prompt-to-image translation, their capacity for reasoning-informed generation remains underexplored. Existing benchmarks focus predominantly on literal text-image alignment (e.g., object existence, spatial arrangements) but fail to evaluate whether models possess reasoning abilities to uncover the deeper meaning behind the text and generate logically coherent visual content. To this end, we identify four dimensions that challenge T2I models to reason about the instructions with domain knowledge before visualizing them:

\noindent \textbf{Idiom Interpretation}. An idiom is a phrase or combination of words with a figurative meaning that differs from its literal meaning. Idioms are common in everyday language, and their meanings usually cannot be deduced by analyzing individual words. For T2I models, prompts containing idioms demand reasoning to obtain the latent meaning before generating correct visual content. This process requires leveraging linguistic knowledge and effectively analyzing context.

\noindent By sourcing from a book~\citep{idiombook} and the internet, we collect 200 idioms that are commonly used in daily life but challenging for T2I models. We then use an LLM
to generate sentences containing the idioms but without explicitly revealing their meanings. These idioms span diverse topics such as social interactions, lifestyle, and emotions. For example, the sentence ``He told a funny joke to break the ice at the start of the meeting'' uses the idiom ``break the ice'', which means to ease tension, rather than literally destroying the ice.

\noindent \textbf{Textual Image Design.} Rich-text images combine visuals and text harmoniously. These images are used to serve specific communicative goals, such as education, marketing, and promotion. Generating such content requires T2I models reasoning about the purpose behind the image and applying goal-oriented design skills like text-visual layout planning, information structuring and rendering. 

\noindent In this dimension, we first collected 200 real-world images featuring rich text from different datasets. Using an MLLM, 
we then extracted the underlying design intentions from these images, resulting in 200 design prompts. Each focuses on the functional purpose of the image rather than describing visual details. For example, ``Design an infographic on online risks for children aged 9-16''. Based on the image sources, the prompts span categories like infographics, posters, documents, and diagrams.

\noindent \textbf{Entity Reasoning.} In everyday life, people often forget specific entity names but remember related details. For example, the prompt ``Generate an image of the team lifting the trophy at the 2022 FIFA World Cup'' requires the T2I model to reason about the context and then retrieve relevant knowledge to generate the entities not explicitly stated. 

\noindent In Entity Reasoning, we begin by defining subdomains for various entities, such as celebrities, artifacts, and architectures. We manually create several example prompts to guide an LLM 
in generating more prompts. After collecting 200 such prompts, we carefully review them to ensure overall consistency and correctness.

\noindent \textbf{Scientific Reasoning.} Creating scientifically realistic images remains a persistent challenge for T2I models, which often produce counterintuitive results that violate scientific laws. 
During pilot generation, we found that unconstrained scientific prompts often allow T2I models to produce images that satisfy the literal object-level description while violating the intended physical scenario. This ambiguity makes scientific evaluation unreliable, because an apparent reasoning failure may instead be caused by under-specification in the prompt. For example, the prompt ``A rubber duck and a metal ball in a water tank'' does not specify the tank size, water condition, or observation state. A model may therefore place the metal ball on top of the duck in an undersized tank, leaving the duck submerged or unable to float, while still following the literal prompt correctly.  
To address this issue, we use more precise prompt to specify the relevant environmental and boundary conditions: ``A rubber duck and a solid steel ball have been placed at separate locations in a large tank full of clear, still water, where the entire system has reached a state of stable equilibrium.'', which clearly state the container, initial configuration, and observation state. These constraints reduce unintended interpretations while preserving the underlying scientific inference.

We organize the prompts into three scientific disciplines: physics, chemistry, and biology. We first manually design representative prompts, use them to guide LLM-based expansion, and then manually verify every prompt for physical correctness, an unambiguous setup, and a non-trivial reasoning requirement.

Figure~\ref{fig:prompt_suite} demonstrates the prompt collection process (left), shows the subcategories in each dimension (middle) and provides the prompt suite statistics (right). 
We visualize the word distribution in Figure~\ref{fig:word_cloud}.
For more information about the prompt suite, please refer to Appendix~\ref{app:prompt}.

\section{Evaluation}
\label{sec:metrics}

\subsection{Evaluation Metric}
\begin{figure*}[!h]
    \includegraphics[width=\linewidth]{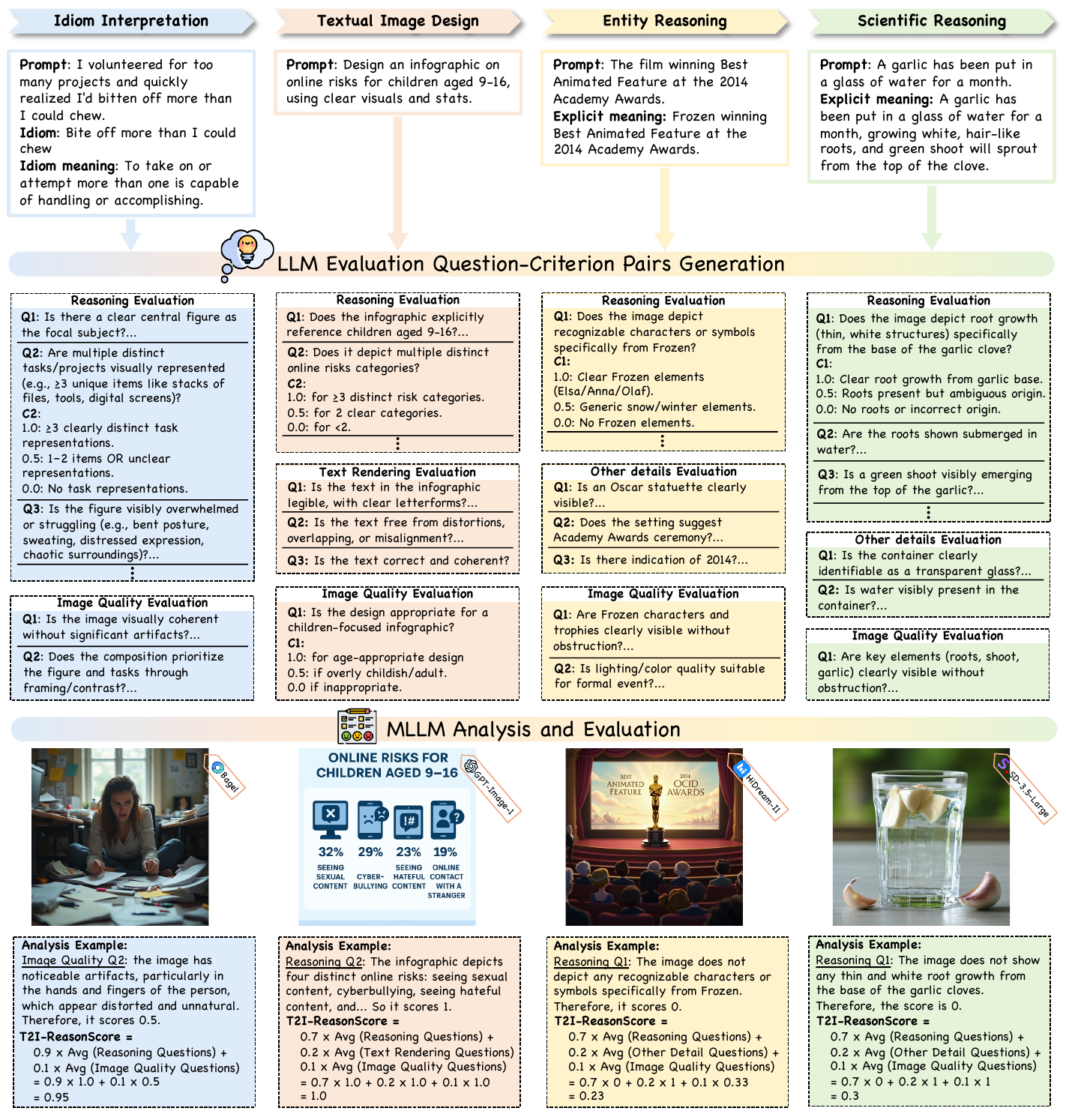}
    \caption{\textbf{Evaluation Framework of T2I-ReasonBench.} We adopt a two-stage evaluation framework: prompt-specific evaluation question-criterion pairs generation by an LLM, then image analysis and scoring by an MLLM. This figure shows one evaluation example for each dimension.}
    \vspace{-14pt}
    \label{fig:metric}
\end{figure*}

In recent years, MLLMs have demonstrated remarkable capabilities in understanding complex visual content, becoming the primary tool for evaluating visual contents.
However, the prompts in our benchmark are highly complex, often involving multiple objects, intricate relationships, and challenging scenarios. As a result, using generic evaluation instructions that are identical for all prompts proves ineffective. This is because each image, generated from a unique prompt, demands specific and targeted checks that generic instructions cannot provide. 
To address this, we develop a two-stage evaluation framework with customized evaluation instructions for each prompt. These instructions take into account the prompt category, the reasoning needed, the explicit content 
the image should exhibit.
Figure~\ref{fig:metric} illustrates the evaluation process.

\noindent \textbf{Prompt-specific question-criterion pairs generation.}
In the first stage, we use an LLM to generate question-criterion pairs based on the prompt and dimension-specific assistant information (e.g., idiom meaning for Idiom Interpretation and explicit meaning for Entity or Scientific Reasoning). For Idiom Interpretation, we generate separate question sets for reasoning accuracy and image quality. For Textual Image Design, we additionally generate a text-rendering question set to disentangle design-intent satisfaction from text quality. For Entity and Scientific Reasoning, where prompts may include explicit details that do not require reasoning, we generate an additional detail-oriented question set to evaluate these details independently. 
Examples are shown in the upper part of Figure~\ref{fig:metric}, with sample questions (Q1, Q2...) and criteria (C1, C2...) in different evaluation sets for each prompt.

\noindent \textbf{Image-level evaluation score.}
In the second stage, we employ an MLLM 
to evaluate the generated images with a Chain-of-Thought~\citep{wei2022chain} (CoT) mechanism: the model first describes the image, then answers the specific questions posed in Stage 1. For each question, the MLLM provides an analysis prior to assigning a score, ensuring thorough and reliable evaluation.
Scores within each set are averaged:
\begin{equation} 
S_{j} = \frac{\sum_{i=1}^{n_j}score_i}{n_j},
\end{equation}
where $S_{j}$ is the score for the $j^{th}$ question set and $n_j$ is the question count for that set. The question sets cover reasoning, detail, and image quality. The final metric, which we call `T2I-ReasonScore', is calculated as:
\begin{multline}
\text{T2I-ReasonScore} = \\
w_1 S_{reason} + w_2 S_{detail} + w_3 S_{quality}
\end{multline}
The lower part of Figure~\ref{fig:metric} illustrates this process to obtain the image-level evaluation score.

In this way, our evaluation metrics reflect the reasoning challenges and provide a comprehensive assessment.
For more details of our evaluation framework, including how we generate the  questions and criteria, please refer to Appendix~\ref{app:evaluation}.

\subsection{Human Evaluation Correlation Analysis} 
To validate the effectiveness of `T2I-ReasonScore', we conduct a human evaluation study to assess its correlation with human judgment. We randomly sample 20 prompts from each of the four dimensions. Using 8 distinct models, we generate 160 images per dimension, yielding a total of 640 images. These images are evaluated by a cohort of postgraduate students based on dimension-specific criteria. Each image is independently scored by 3 annotators, and the results are averaged to establish a gold-standard human score. Then we calculate the correlation between automatic metrics and the human scores using Kendall's $\tau$ and Spearman's $\rho$.

\noindent To determine the most effective model configuration for our metric, we compute T2I-ReasonScore using 12 distinct LLM-MLLM combinations.
For question-criterion generation, we employ three different LLMs: DeepSeek-R1~\citep{guo2025deepseek}, GPT-5.1~\citep{GPT-5.1}, and Gemini-2.5-pro~\citep{Gemini-2.5-pro}. For image evaluation, we use five MLLMs: Qwen2.5-VL-72B~\citep{Qwen2.5-VL}, GPT-5.1, LLaVA-OneVision-1.5~\citep{an2025llava}, Gemini-2.5-pro, and Qwen3-VL-235B-A22B~\citep{bai2025qwen3}. Finally, to assess the robustness of our two-stage evaluation pipeline, we compare our metric against two widely-used T2I metrics: CLIPscore~\citep{hessel2021clipscore} and VQAscore~\citep{lin2024evaluating}.

The correlation results, presented in Table~\ref{tab:human_corr_select}, reveal two key findings: first, the proposed metric consistently outperforms traditional text-image alignment metrics (CLIPscore and VQAscore). Second, Gemini-2.5-pro proves to be a stable and high-performing MLLM evaluator, provided it is paired with a capable LLM in the preceding stage.

The average correlation is the highest overall when pairing Gemini-2.5-pro with questions from GPT-5.1. However, in Entity Reasoning, Gemini-2.5-pro with questions generated by itself yields higher correlation. Therefore, we select Gemini-2.5-pro as our unified MLLM evaluator and employ a hybrid strategy for question generation: Gemini-2.5-pro for Entity Reasoning and GPT-5.1 for the other three dimensions.
Figure~\ref{fig:qualitative} in Appendix~\ref{app:qualitative} shows more qualitative examples generated by different T2I models with the evaluated T2I-ReasonScore. It can be seen that the scores are highly aligned with human preference. For more details about human evaluation, please refer to Appendix~\ref{app:evaluation}.

\begin{table}[t]
\centering
    \caption{\textbf{Correlation between automatic metrics and human evaluation.} We use different LLM-MLLM combinations to calculate our metric `T2I-ReasonScore'. For each dimension, \textbf{Bold} signifies the highest correlation, \underline{underline} denotes the second highest. 
    We highlight the combination adopted for each dimension in \colorbox{mycolor_yellow}{yellow}.
    } 
    \vspace{-5pt}

\resizebox{\linewidth}{!}{%  
\begin{tabular}{ll cc cc cc cc cc}
\toprule
\multirow{2}{*}{\textbf{LLM}} & \multirow{2}{*}{\textbf{MLLM}} & \multicolumn{2}{c}{\textbf{Idiom}} & \multicolumn{2}{c}{\textbf{Textual}} & \multicolumn{2}{c}{\textbf{Entity}} & \multicolumn{2}{c}{\textbf{Scientific}} & \multicolumn{2}{c}{\textbf{Average}} \\

\cmidrule(lr){3-4}\cmidrule(lr){5-6}\cmidrule(lr){7-8}\cmidrule(lr){9-10}\cmidrule(lr){11-12}
 & & $\tau$($\uparrow$) & $\rho$($\uparrow$) & $\tau$($\uparrow$) & $\rho$($\uparrow$) & $\tau$($\uparrow$) & $\rho$($\uparrow$) & $\tau$($\uparrow$) & $\rho$($\uparrow$) & $\tau$($\uparrow$) & $\rho$($\uparrow$)    \\
    \midrule
\multicolumn{2}{l}{\hspace{-.5em} \textit{T2I-ReasonScore}} \\
Deepseek & Qwen2.5-VL           & 0.5095  &  0.6792  &  0.6051  &  0.7686  &  0.4767  &  0.6242  &  0.4984  &  0.6438  &  0.5224  &  0.6790  \\
Deepseek & GPT-5.1              & 0.4943  &  0.6659  &  0.6052  &  0.7493  &  0.5432  &  0.6831  &  0.4704  &  0.6168  &  0.5283  &  0.6788  \\
Deepseek & LLaVA-1V             & 0.4664  &  0.6202  &  0.5561  &  0.6936  &  0.5116  &  0.6377  &  0.4342  &  0.5802  &  0.4921  &  0.6329  \\    
\midrule
GPT-5.1 & Qwen2.5-VL            & 0.5391  &  0.7004  &  0.6372  &  0.8002  &  0.5937  &  0.7431  &  0.5062  &  0.6592  &  0.5691  &  0.7257  \\
GPT-5.1 & GPT-5.1               & \underline{0.5576}  &  \underline{0.7168}  &  0.6446  &  \underline{0.8076}  &  0.6077  &  0.7642  &  0.4589  &  0.6098  &  0.5672  &  0.7246  \\
GPT-5.1 & LLaVA-1V              & 0.5165  &  0.6594  &  0.5343  &  0.6619  &  0.5762  &  0.7074  &  0.3952  &  0.5130  &  0.5055  &  0.6354  \\
GPT-5.1 & Gemini-2.5-pro        & \colorbox{mycolor_yellow}{0.5540}  &  \colorbox{mycolor_yellow}{0.7115}  &  \colorbox{mycolor_yellow}{0.6175}  &  \colorbox{mycolor_yellow}{0.7848}  &  0.6102  &  0.7605  &  \colorbox{mycolor_yellow}{\textbf{0.5673}}  &  \colorbox{mycolor_yellow}{\textbf{0.7231}}  &  \textbf{0.5872}  &  \textbf{0.7450}  \\
GPT-5.1 & Qwen3-VL              & \textbf{0.5596}  &  \textbf{0.7257}  &  0.6324  &  0.7913  &  0.5611  &  0.7114  &  \underline{0.5285}  &  \underline{0.6778}  &  0.5704  &  0.7265  \\
\midrule
Gemini-2.5-pro & Qwen2.5-VL     & 0.4822  &  0.6320  &  \textbf{0.6829}  &  \textbf{0.8224}  &  0.6344  &  \underline{0.7882}  &  0.4245  &  0.5641  &  0.5560  &  0.7017  \\
Gemini-2.5-pro & GPT-5.1        & 0.5218  &  0.6672  &  0.6458  &  0.7896  &  \textbf{0.6514}  &  \textbf{0.7968}  &  0.4343  &  0.5781  &  0.5633  &  0.7079  \\
Gemini-2.5-pro & Gemini-2.5-pro & 0.4975  &  0.6566  &  0.6096  &  0.7641  &  \colorbox{mycolor_yellow}{\underline{0.6390}}  &  \colorbox{mycolor_yellow}{0.7830}  &  0.4471  &  0.5918  &  0.5483  &  0.6989  \\
Gemini-2.5-pro & Qwen3-VL       & 0.4693  &  0.6132  &  \underline{0.6527}  &  0.7947  &  0.6071  &  0.7399  &  0.5109  &  0.6545  &  0.5600  &  0.7006  \\

\midrule
\multicolumn{2}{l}{\hspace{-.5em} \textit{Other metrics}} \\
\multicolumn{2}{l}{CLIPscore~\cite{hessel2021clipscore} }&$0.3186$&$0.4348$ &$0.5372$&$0.7187$ &$0.2732$&$0.3837$ &$0.1905$&$0.2657$&$0.3299$&$0.4507$ \\
\multicolumn{2}{l}{VQAscore~\cite{lin2024evaluating}}&$0.4091$&$0.5672$ &$0.4890$&$0.6590$ &$0.4483$&$0.6133$ &$0.3698$&$0.4939$ &$0.4291$&$0.5834$ \\
\bottomrule
\end{tabular}
}
\label{tab:human_corr_select}
\vspace{-10pt}
\end{table}
\section{Evaluation Results}
\label{sec:experiment}

\textbf{Evaluated models.}
We evaluate 16 state-of-the-art T2I models, including 8 diffusion models, 5 unified multimodal models, and 3 proprietary models. The diffusion models are HiDream-I1-full~\citep{hidream}, FLUX.1-dev~\citep{flux2024}, FLUX.1-schnell~\citep{flux2024}, Playground-v2.5~\citep{playground}, Stable-Diffusion-3-Medium~\citep{sd3medium}, Stable-Diffusion-3.5-Medium~\citep{sd3medium}, Stable-Diffusion-3.5-Large~\citep{sd3medium}, Qwen-Image~\citep{wu2025qwen}.
Unified models: 
Bagel~\citep{deng2025emerging}, Emu3~\citep{emu3}, Janus-Pro-7B~\citep{janus}, show-o-demo-512~\citep{showo}, GoT~\citep{got}.
Proprietary models are Seedream 4.0~\citep{seedream2025}, GPT-Image-1~\citep{GPT}, Nano-Banana~\citep{nano-banana}.

\noindent \textbf{Implementation Details.} We use the default settings for all T2I models in image generation. We set the weights $[w_1, w_2, w_3]$ in T2I-ReasonScore to $[0.9, 0.0, 0.1]$ for Idiom Interpretation, and $[0.7, 0.2, 0.1]$ for Textual Image Design, Entity and Scientific Reasoning to prioritize reasoning while maintaining a balanced final score.  

\begin{figure}[!h]
    \includegraphics[width=\linewidth]{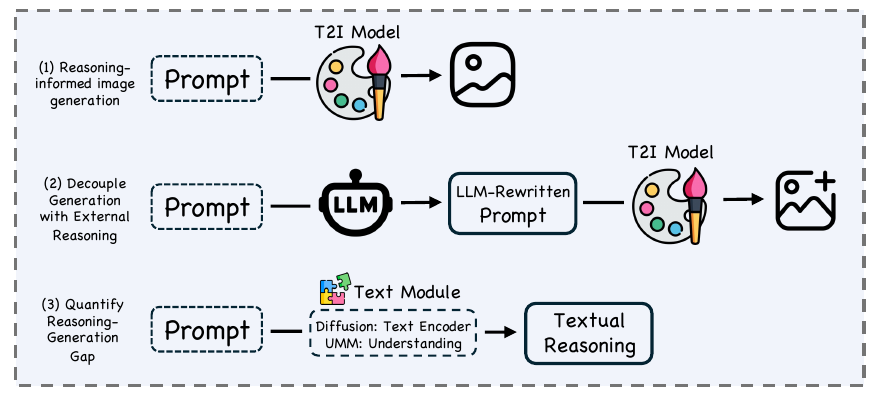}
    \caption{\textbf{Overview of the Experiment Settings.}
The first experiment (\S\ref{sec:experiment}.1) establishes a baseline, where images are generated directly from the original benchmark prompts. The second (\S\ref{sec:experiment}.2) introduces an external LLM to reason and rewrite the prompts before they are sent to the T2I models for image generation. The third experiment (\S\ref{sec:experiment}.3) isolates and evaluates the models' internal reasoning by using their text encoders or understanding mode to generate textual reasoning, the quality of which is measured as a text score.}
    \vspace{-10pt}
    \label{fig:analysis}
\end{figure}

\begin{table}[t]
\centering
\caption{\textbf{Evaluation results of T2I-ReasonBench.} Scores are normalized between 0-100. A higher score indicates better performance. \colorbox{mycolor_blue}{Blue} highlights the top score in diffusion models. \colorbox{mycolor_yellow}{Yellow} highlights the top score in unified multimodal models. \textbf{Bold} signifies the highest score of all models. 
}
\vspace{2pt}
\label{tab:result}
\resizebox{\linewidth}{!}{%

\begin{tabular}{l|cccc|c} 
\toprule
\multicolumn{1}{c}{Model} & Idiom & Textual & Entity & Scientific & Overall \\

\midrule
 & \multicolumn{5}{c}{\textbf{Diffusion Models}} \\
\midrule
SD-3-Medium~\cite{sd3medium} & 41.7 & 69.3 & 43.1 & 53.8 & 52.0 \\
SD-3.5-Medium~\cite{sd3medium} & 39.6 & 68.4 & 47.6 & 54.1 & 52.4 \\
SD-3.5-Large~\cite{sd3medium} & 44.2 & 68.8 & 50.9 & 58.6 & 55.6 \\
FLUX.1-dev~\cite{flux2024} & 49.3 & 72.1 & 50.5 & 56.3 & 57.1 \\
FLUX.1-schnell~\cite{flux2024} & 47.4 & 78.3 & 50.1 & 61.1 & 59.2 \\
Playground-v2.5~\cite{playground} & 50.8 & 51.2 & 51.1 & 50.7 & 50.9 \\
HiDream-I1-full~\cite{hidream} & 59.1 & \colorbox{mycolor_blue}{81.8} & 58.0 & 59.9 & 64.7 \\
Qwen-Image~\cite{wu2025qwen} & \colorbox{mycolor_blue}{62.7} & 77.6 & \colorbox{mycolor_blue}{61.8} & \colorbox{mycolor_blue}{69.0} & \colorbox{mycolor_blue}{67.8} \\
\midrule
 & \multicolumn{5}{c}{\textbf{Unified Multimodal Models}} \\
\midrule
Emu3~\cite{emu3} & 39.2 & 44.6 & 42.7 & 42.9 & 42.4 \\
show-o-demo-512~\cite{showo} & 38.9 & 47.2 & 41.0 & 49.2 & 44.1 \\
Janus-Pro-7B~\cite{janus} & 32.7 & 45.8 & 41.3 & 44.4 & 41.0 \\
GoT~\cite{got} & 33.7 & 48.9 & 37.6 & 42.8 & 40.7 \\
Bagel w/ Thinking~\cite{deng2025emerging} & \colorbox{mycolor_yellow}{50.6} & \colorbox{mycolor_yellow}{56.8} & \colorbox{mycolor_yellow}{56.5} & \colorbox{mycolor_yellow}{66.5} & \colorbox{mycolor_yellow}{57.6} \\
\midrule
& \multicolumn{5}{c}{\textbf{Proprietary Models}} \\
\midrule
Seedream 4.0~\cite{seedream2025} & 84.8 & 94.1 & 73.0 & 84.4 & 84.0 \\
GPT-Image-1~\cite{GPT} & 84.1 & \textbf{94.4} & 85.7 & 87.1 & 87.8 \\
Nano-Banana~\cite{nano-banana} & \textbf{89.8} & 94.0 & \textbf{87.3} & \textbf{89.9} & \textbf{90.3} \\
\bottomrule   
\end{tabular}
}
\vspace{-10pt}
\end{table}

\subsection{Quantitative Evaluation}
The evaluation results on T2I-ReasonBench, presented in Table~\ref{tab:result}, reveal significant limitations and pronounced performance gaps in current T2I models. Proprietary models demonstrate a clear lead, with scores consistently exceeding 70 points and Nano-Banana approaching 90. In contrast, open-source models lag far behind: specialized diffusion models typically score around 50; unified multimodal models are even lower, with most around 40 points, though Bagel is a notable exception, performing on par with the diffusion models.

\subsection{Decoupling Generation via External Reasoning}
We conduct an additional experiment using a pipeline that decouples image generation from reasoning, as shown in (2) of Figure~\ref{fig:analysis}. In this setup, GPT-4o~\citep{gpt4o} first reasons about the original prompt and converts it into a visually explicit description, which is then fed to the T2I model. Table~\ref{tab:result_pipeline} presents the quantitative results of using the LLM-rewritten prompts.

\begin{table}[t]
\centering
\caption{\textbf{Evaluation results with LLM-rewritten prompts.} 
\textcolor{RedOrange}{$\uparrow$ Red} and \textcolor{BlueGreen}{$\downarrow$ Blue} subscripts denote score changes relative to original prompts. 
}
\label{tab:result_pipeline}
\resizebox{\linewidth}{!}{%

\begin{tabular}{l|cccc|c} 
\toprule
\multicolumn{1}{c}{Model} & Idiom & Textual & Entity & Scientific & Overall \\

\midrule
 & \multicolumn{5}{c}{\textbf{Diffusion Models}} \\
\midrule
SD-3-Medium & 76.7\red{34.9} & 76.0\red{6.8} & 75.0\red{32.0} & 68.7\red{14.9} & 74.1\red{22.1} \\
SD-3.5-Medium & 76.4\red{36.8} & 75.6\red{7.2} & 76.4\red{28.8} & 68.7\red{14.5} & 74.3\red{21.8} \\
SD-3.5-Large & 76.7\red{32.5} & 75.8\red{7.0} & 82.1\red{31.1} & 72.6\red{14.0} & 76.8\red{21.2} \\
FLUX.1-dev & 77.8\red{28.5} & 80.3\red{8.2} & 75.6\red{25.1} & 74.0\red{17.7} & 76.9\red{19.9} \\
FLUX.1-schnell & 79.9\red{32.5} & 80.5\red{2.1} & 76.6\red{26.5} & 74.9\red{13.8} & 78.0\red{18.7} \\
Playground-v2.5 & 65.3\red{14.5} & 53.5\red{2.3} & 77.4\red{26.4} & 55.6\red{4.9} & 62.9\red{12.0} \\
HiDream-I1-full & 77.0\red{17.9} & \colorbox{mycolor_blue}{84.6}\red{2.8} & 79.6\red{21.6} & 71.0\red{11.1} & 78.1\red{13.4} \\
Qwen-Image & \colorbox{mycolor_blue}{84.7}\red{22.0} & 82.7\red{5.0} & \colorbox{mycolor_blue}{84.2}\red{22.4} & \colorbox{mycolor_blue}{81.4}\red{12.4} & \colorbox{mycolor_blue}{83.3}\red{15.5} \\
\midrule
 & \multicolumn{5}{c}{\textbf{Unified Multimodal Models}} \\
\midrule
Emu3 & 67.7\red{28.5} & 52.4\red{7.7} & 67.6\red{24.9} & 54.2\red{11.3} & 60.5\red{18.1} \\
show-o-demo-512 & 74.8\red{35.8} & 51.2\red{4.0} & 69.4\red{28.4} & 64.4\red{15.1} & 64.9\red{20.9} \\
Janus-Pro-7B & 72.6\red{39.9} & 57.4\red{11.7} & 71.0\red{29.7} & 64.9\red{20.5} & 66.5\red{25.5} \\
GoT & 62.3\red{28.7} & 52.7\red{3.8} & 55.2\red{17.6} & 47.6\red{4.7} & 54.5\red{13.7} \\
Bagel w/o Thinking & \colorbox{mycolor_yellow}{77.9}\red{27.4} & \colorbox{mycolor_yellow}{71.4}\red{14.6} & \colorbox{mycolor_yellow}{73.9}\red{17.5} & \colorbox{mycolor_yellow}{75.1}\red{8.6} & \colorbox{mycolor_yellow}{74.6}\red{17.0} \\
\midrule
& \multicolumn{5}{c}{\textbf{Proprietary Models}} \\
\midrule
Seedream 4.0 & 87.3\red{2.5} & 87.1\blue{7.0} & 86.8\red{13.8} & 87.9\red{3.5} & 87.3\red{3.2} \\
GPT-Image-1 & 87.1\red{3.1} & 92.3\blue{2.2} & \textbf{90.8}\red{5.0} & \textbf{89.3}\red{2.2} & 89.9\red{2.0} \\
Nano-Banana & \textbf{87.7}\blue{2.2} & \textbf{92.7}\blue{1.3} & 90.2\red{2.9} & 89.1\blue{0.8} & \textbf{89.9}\blue{0.4} \\
\bottomrule   
\end{tabular}
}
\vspace{-10pt}
\end{table}

\noindent \textbf{Reasoning is the bottleneck for model performance.}
By providing models with pre-reasoned, visually explicit prompts, this pipeline setting allows us to isolate and clarify their actual T2I generation capability. The results show that this approach substantially improves the accuracy of reasoning for open-source models. \textbf{\textit{The more a model improves, the more it initially lacks reasoning ability, as opposed to generation ability.}}
As a result, the performance gap between open-source and proprietary models shrinks with decoupled reasoning, suggesting that the previous disparity in Table~\ref{tab:result} stems primarily from differences in reasoning ability. A component-level breakdown for both original and LLM-rewritten prompts is provided in Appendix~\ref{app:component_scores}.

\subsection{Quantifying the Internal Reasoning-Generation Gap}
We design an experiment to measure the reasoning capability of each model's text module and quantify the internal reasoning-generation gap. It involves isolating the text understanding component and having it generate a textual description with concrete visual details for each prompt, as illustrated in (3) of Figure~\ref{fig:analysis}. This textual output is then evaluated using our metric (without evaluating the image quality) to produce a ``text score'', as shown in Table~\ref{tab:text_result}. 
This score reflects the pure reasoning ability of the text module, representing a theoretical upper limit on the model's performance if its reasoning were perfectly utilized by generation. The difference between this text score and the model's final image score provides an approximation of its reasoning-generation gap.

\begin{table}[t]
\centering
\caption{\textbf{Comparison of Reasoning Ability in Image Generation vs. Text-Only Interpretation.}
Scores are presented as \textit{Image Score/Text Score}. The \textit{Text Score} is derived by evaluating an explicit description generated by the model's isolated text module. The \textit{Image Score} evaluates the image generated by the full model. Both scores are composed of $w_1 S_{reason} + w_2 S_{detail}$, which excludes the image quality component, for a maximum possible value of 90. \textbf{Bold} marks the lowest scores.}

\vspace{-5pt}
\label{tab:text_result}
\resizebox{\linewidth}{!}{%

\begin{tabular}{l|cccc} 
\toprule
\multicolumn{1}{c}{Model/Text-Module} & Idiom & Textual & Entity & Scientific \\

\midrule
\multicolumn{2}{l}{\hspace{-.5em} \textit{Diffusion Models}} \\
HiDream-I1-full/Llama-3.1-8b     & 50.3/74.1 & 71.2/81.5 & 49.1/66.8 & 50.9/69.6 \\                            
Qwen-Image/Qwen2.5-VL-7b         & 53.1/70.2 & 69.2/85.1 & 52.5/63.0 & 59.3/70.5 \\ 
\midrule
\multicolumn{2}{l}{\hspace{-.5em} \textit{Unified Multimodal Models}} \\
Emu3-Gen/Emu3-Und                & \textbf{31.3/36.9} & \textbf{34.9/80.3} & \textbf{34.9/56.9} & \textbf{37.7/63.2} \\                                        
Bage-Gen/Bagel-Und    & 41.7/52.0 & 43.2/84.4 & 47.8/67.6 & 56.0/76.4 \\         
\midrule  
\multicolumn{2}{l}{\hspace{-.5em} \textit{Proprietary Models}} \\
GPT-Image-1/GPT-4o               & 74.3/78.2 & 82.7/87.6 & 76.0/73.6 & 73.2/77.4 \\                                               
Nano-Banana-Gen/Nano-Banana-Und  & 80.1/76.7 & 85.5/86.8 & 77.7/75.1 & 76.2/80.9 \\                                                                                  
\bottomrule   
\end{tabular}
}
\vspace{-5pt}
\end{table}

\noindent \textbf{Proprietary models demonstrate superior reasoning ability and minimal reasoning-generation gap.}
Proprietary models achieve the highest text reasoning scores, representing their superior ability to understand reasoning-intensive prompts. Nano-Banana~\citep{nano-banana}, for example, verbalizes its reasoning before generation. For the prompt ``The city hosting the Summer Olympics in 2021'', it explicitly identifies Tokyo and then plans to include its landmarks and Olympic imagery.
While other top models like GPT-image-1 do not provide such textual response, the minimal gap between their text scores and image scores demonstrate a highly efficient transfer of reasoning to synthesis. 
Our experiment further validates this conclusion. As shown in Figure~\ref{fig:good_quality}, there is a high degree of alignment between the final images in the third column and the correct textual reasoning that we prompt the models to generate in a separate query. It demonstrates that the models' internal reasoning is stable and that the generation process fully integrates this reasoning, reflecting it in the final image without information loss.

\noindent \textbf{Leading diffusion models still limited by semantic mapping.}
The cases of Qwen-Image~\citep{wu2025qwen} and HiDream~\citep{hidream} reveal a critical reasoning-generation gap in top-tier diffusion models. Both incorporate powerful LLMs as text encoders (Qwen2.5-VL-7b and Llama 3.1-8b), whose reasoning abilities are comparable to proprietary models. However, this advantage is lost during image synthesis.
As shown in Figure~\ref{fig:bad_quality} (Row\#1,\#2), while the models are able to generate correct imagery from LLM pre-reasoned prompts, they fail with the original implicit prompts. In the latter case, even though the isolated text encoders can correctly interpret them, the generated images only reflect a superficial keyword mapping, leading to incorrect outputs. This demonstrates that the models fail to leverage the deep reasoning capabilities of their LLM encoders. Instead, they function merely as semantic mappers rather than integrated reasoners, which ultimately limits the models' performance on reasoning-aware image generation.

\noindent \textbf{Unified multimodal models face both reasoning and generation challenges.}
Early UMM like Emu3 struggles with both reasoning and generation. As shown in Figure~\ref{fig:bad_quality} (Row \#4). Rather than analyzing the scientific implications, its textual output merely extends the original prompt with more descriptive details. This reasoning deficit is compounded by poor generation quality, as the model fails to compose all objects specified in the prompt.

\noindent More advanced UMM like Bagel exhibits stronger reasoning capabilities. It unifies LLM and diffusion model within a single transformer architecture, enhancing interaction between understanding and generation modules. It can  analyze a prompt's implicit meaning and verbalize its reasoning before generation (Figure~\ref{fig:bad_quality}, Row \#3). However, even when its textual reasoning is correct, the final generated image may fail to align with that plan, indicating a failure to transfer its understanding into the visual synthesis.

\noindent \textbf{Insights for future model design.}
The last two rows in Figure~\ref{fig:bad_quality} highlight two distinct failure cases in proprietary models: one of reasoning and one of integration. The first is a reasoning failure (Row \#5), where Nano-Banana's textual analysis fails to uncover the prompt's physical implications, leading directly to an incorrect image. The second is a reasoning-integration failure (Row \#6). Here, GPT-4o's textual reasoning correctly recognizes the entity, yet the model neglects the implicit meaning when generating the image.
\noindent 
These examples highlight a key challenge: improving reasoning-aware text-to-image generation requires not only stronger foundational reasoning in current T2I models, but also more effective integration of that reasoning into the final image synthesis.

\section{Conclusion}
We introduce T2I-ReasonBench, a benchmark for evaluating reasoning-informed text-to-image generation. Unlike prior benchmarks that mainly emphasize literal prompt following, T2I-ReasonBench focuses on whether models can infer implicit meaning, apply world knowledge, and translate reasoning outcomes into coherent visual outputs. To support this goal, we curate 900 prompts across four challenging dimensions: Idiom Interpretation, Textual Image Design, Entity Reasoning, and Scientific Reasoning, and propose a prompt-specific two-stage evaluation framework with T2I-ReasonScore for fine-grained and interpretable assessment.

Through a comprehensive study of 16 state-of-the-art models, we show that current T2I systems still face substantial limitations in reasoning-aware generation. Our results reveal two major bottlenecks: first, many models struggle to correctly reason about prompts containing implicit meaning; second, even when a model's text understanding module is capable of correct reasoning, this understanding is often not faithfully reflected in the final generated image. 

Overall, T2I-ReasonBench highlights that high image quality and literal text alignment are not sufficient to measure true text-to-image intelligence. Future progress in T2I generation will require both stronger foundational reasoning and tighter integration between reasoning and visual synthesis. We hope this benchmark can serve as a useful testbed for the community, encouraging the development of next-generation T2I systems that not only generate realistic images, but also genuinely understand what they are asked to depict.

\section{Societal Impacts}
This work may support the development of more reliable reasoning-aware text-to-image models, which could benefit applications such as education, design, and scientific communication. At the same time, stronger reasoning ability may also enable the generation of more convincing misleading or deceptive images, including fake advertisements, documents, or propaganda.
The benchmark itself may also introduce risks. By emphasizing reasoning performance, it could incentivize building more capable generative systems without corresponding safeguards. 
Overall, we view T2I-ReasonBench as an evaluation tool, and we believe progress on reasoning-aware generation should be accompanied by appropriate safeguards.

% \clearpage
\section*{Limitations}
This work has several limitations. 

    First, although T2I-ReasonBench covers four important reasoning dimensions, it does not exhaust the full spectrum of reasoning required in T2I generation. Other forms of reasoning could be explored in future extensions.

    Second, while we design a prompt-specific two-stage evaluation framework and show that T2I-ReasonScore correlates well with human judgments, the evaluation still relies on LLMs and MLLMs as automatic judges. Such evaluators may themselves introduce bias, make recognition errors, or favor certain visual styles and model outputs. Although human correlation analysis supports the reliability of our metric, automatic evaluation cannot fully replace large-scale human assessment.

    Third, the benchmark is currently limited in scale and language coverage. Our prompt set contains 900 English prompts, which enables controlled and diverse evaluation, but may not fully represent the richness of real-world user instructions, multilingual usage, or culturally dependent expressions such as region-specific idioms and design conventions. Extending the benchmark to broader languages and cultural contexts would improve its generality.

    Finally, benchmark construction and prompt verification involve LLM assistance and expert review, which may still leave room for subjective choices in prompt design, and expected visual interpretation. While we carefully validate prompts to ensure quality and consistency, some cases may still admit multiple plausible visual realizations. We hope future work can further expand the benchmark, diversify the reasoning dimensions, and develop even more robust evaluation protocols.

\bibliography{custom}

@inproceedings{hu2023tifa,
  title={Tifa: Accurate and interpretable text-to-image faithfulness evaluation with question answering},
  author={Hu, Yushi and Liu, Benlin and Kasai, Jungo and Wang, Yizhong and Ostendorf, Mari and Krishna, Ranjay and Smith, Noah A},
  booktitle={Proceedings of the IEEE/CVF International Conference on Computer Vision},
  pages={20406--20417},
  year={2023}
}

@article{huang2023t2icompbench,
  title={T2I-CompBench: A Comprehensive Benchmark for Open-world Compositional Text-to-Image Generation},
  author={Huang, Kaiyi and Sun, Kaiyue and Xie, Enze and Li, Zhenguo and Liu, Xihui},
  journal={Advances in Neural Information Processing Systems},
  volume={36},
  pages={78723--78747},
  year={2023}
}

@article{ghosh2023geneval,
  title={GenEval: An Object-Focused Framework for Evaluating Text-to-Image Alignment},
  author={Ghosh, Dhruba and Hajishirzi, Hannaneh and Schmidt, Ludwig},
  journal={arXiv preprint arXiv:2310.11513},
  year={2023}
}

@article{hu2024ella,
  title={Ella: Equip diffusion models with llm for enhanced semantic alignment},
  author={Hu, Xiwei and Wang, Rui and Fang, Yixiao and Fu, Bin and Cheng, Pei and Yu, Gang},
  journal={arXiv preprint arXiv:2403.05135},
  year={2024}
}

@article{li2024genai,
  title={Genai-bench: Evaluating and improving compositional text-to-visual generation},
  author={Li, Baiqi and Lin, Zhiqiu and Pathak, Deepak and Li, Jiayao and Fei, Yixin and Wu, Kewen and Ling, Tiffany and Xia, Xide and Zhang, Pengchuan and Neubig, Graham and others},
  journal={arXiv preprint arXiv:2406.13743},
  year={2024}
}

@article{wu2024conceptmix,
  title={Conceptmix: A compositional image generation benchmark with controllable difficulty},
  author={Wu, Xindi and Yu, Dingli and Huang, Yangsibo and Russakovsky, Olga and Arora, Sanjeev},
  journal={Advances in Neural Information Processing Systems},
  volume={37},
  pages={86004--86047},
  year={2024}
}

@article{huang2025t2icompbench++,
author={Huang, Kaiyi and Duan, Chengqi and Sun, Kaiyue and Xie, Enze and Li, Zhenguo and Liu, Xihui},
journal={ IEEE Transactions on Pattern Analysis \& Machine Intelligence },
title={{ T2I-CompBench++: An Enhanced and Comprehensive Benchmark for Compositional Text-to-Image Generation }},
year={2025},
volume={47},
number={05},
ISSN={1939-3539},
pages={3563-3579},
doi={10.1109/TPAMI.2025.3531907},
url = {https://doi.ieeecomputersociety.org/10.1109/TPAMI.2025.3531907},
publisher={IEEE Computer Society},
address={Los Alamitos, CA, USA},
month=may}

@article{wei2025tiif,
  title={TIIF-Bench: How Does Your T2I Model Follow Your Instructions?},
  author={Wei, Xinyu and Zhang, Jinrui and Wang, Zeqing and Wei, Hongyang and Guo, Zhen and Zhang, Lei},
  journal={arXiv preprint arXiv:2506.02161},
  year={2025}
}

@article{niu2025wise,
  title={WISE: A World Knowledge-Informed Semantic Evaluation for Text-to-Image Generation},
  author={Niu, Yuwei and Ning, Munan and Zheng, Mengren and Lin, Bin and Jin, Peng and Liao, Jiaqi and Ning, Kunpeng and Zhu, Bin and Yuan, Li},
  journal={arXiv preprint arXiv:2503.07265},
  year={2025}
}

@article{meng2024phybench,
  title={PhyBench: A Physical Commonsense Benchmark for Evaluating Text-to-Image Models},
  author={Meng, Fanqing and Shao, Wenqi and Luo, Lixin and Wang, Yahong and Chen, Yiran and Lu, Quanfeng and Yang, Yue and Yang, Tianshuo and Zhang, Kaipeng and Qiao, Yu and Luo, Ping},
  journal={arXiv preprint arXiv:2406.11802},
  year={2024}
}

@article{fu2024commonsense,
  title={Commonsense-T2I Challenge: Can Text-to-Image Generation Models Understand Commonsense?},
  author={Fu, Xingyu and He, Muyu and Lu, Yujie and Wang, William Yang and Roth, Dan},
  journal={arXiv preprint arXiv:2406.07546},
  year={2024}
}

@article{chen2025r2i,
  title={R2I-Bench: Benchmarking Reasoning-Driven Text-to-Image Generation},
  author={Chen, Kaijie and Lin, Zihao and Xu, Zhiyang and Shen, Ying and Yao, Yuguang and Rimchala, Joy and Zhang, Jiaxin and Huang, Lifu},
  journal={arXiv preprint arXiv:2505.23493},
  year={2025}
}

@article{nichol2021glide,
  title={Glide: Towards photorealistic image generation and editing with text-guided diffusion models},
  author={Nichol, Alex and Dhariwal, Prafulla and Ramesh, Aditya and Shyam, Pranav and Mishkin, Pamela and McGrew, Bob and Sutskever, Ilya and Chen, Mark},
  journal={arXiv preprint arXiv:2112.10741},
  year={2021}
}

@article{liu2024lumina,
  title={Lumina-mGPT: Illuminate Flexible Photorealistic Text-to-Image Generation with Multimodal Generative Pretraining},
  author={Liu, Dongyang and Zhao, Shitian and Zhuo, Le and Lin, Weifeng and Qiao, Yu and Li, Hongsheng and Gao, Peng},
  journal={arXiv preprint arXiv:2408.02657},
  year={2024}
}

@article{sun2024autoregressive,
  title={Autoregressive model beats diffusion: Llama for scalable image generation},
  author={Sun, Peize and Jiang, Yi and Chen, Shoufa and Zhang, Shilong and Peng, Bingyue and Luo, Ping and Yuan, Zehuan},
  journal={arXiv preprint arXiv:2406.06525},
  year={2024}
}

@article{chameleon2024mixed,
  title={Chameleon: Mixed-Modal Early-Fusion Foundation Models},
  author={Chameleon Team},
  journal={arXiv preprint arXiv:2405.09818},
  year={2024}
}

@article{ramesh2021zero,
  title={Zero-Shot Text-to-Image Generation},
  author={Ramesh, Aditya and Pavlov, Mikhail and Goh, Gabriel and Gray, Scott and Voss, Chelsea and Radford, Alec and Chen, Mark and Sutskever, Ilya},
  journal={arXiv preprint arXiv:2102.12092},
  year={2021}
}

@article{ding2021cogview,
  title={CogView: Mastering Text-to-Image Generation via Transformers},
  author={Ding, Ming and Yang, Zhuoyi and Hong, Wenyi and Zheng, Wendi and Zhou, Chang and Yin, Da and Lin, Junyang and Zou, Xu and Shao, Zhou and Yang, Hongxia and Tang, Jie},
  journal={arXiv preprint arXiv:2105.13290},
  year={2021}
}

@article{saharia2022photorealistic,
  title={Photorealistic text-to-image diffusion models with deep language understanding},
  author={Saharia, Chitwan and Chan, William and Saxena, Saurabh and Li, Lala and Whang, Jay and Denton, Emily L and Ghasemipour, Kamyar and Gontijo Lopes, Raphael and Karagol Ayan, Burcu and Salimans, Tim and others},
  journal={Advances in Neural Information Processing Systems},
  volume={35},
  pages={36479--36494},
  year={2022}
}

@article{dhariwal2021diffusion,
  title={Diffusion models beat gans on image synthesis},
  author={Dhariwal, Prafulla and Nichol, Alexander},
  journal={Advances in neural information processing systems},
  volume={34},
  pages={8780--8794},
  year={2021}
}

@article{ho2020denoising,
  title={Denoising diffusion probabilistic models},
  author={Ho, Jonathan and Jain, Ajay and Abbeel, Pieter},
  journal={Advances in neural information processing systems},
  volume={33},
  pages={6840--6851},
  year={2020}
}

@article{yu2022scaling,
  title={Scaling autoregressive models for content-rich text-to-image generation},
  author={Yu, Jiahui and Xu, Yuanzhong and Koh, Jing Yu and Luong, Thang and Baid, Gunjan and Wang, Zirui and Vasudevan, Vijay and Ku, Alexander and Yang, Yinfei and Ayan, Burcu Karagol and others},
  journal={arXiv preprint arXiv:2206.10789},
  volume={2},
  number={3},
  pages={5},
  year={2022}
}

@article{vq2,
  title={What you see is what you read? improving text-image alignment evaluation},
  author={Yarom, Michal and Bitton, Yonatan and Changpinyo, Soravit and Aharoni, Roee and Herzig, Jonathan and Lang, Oran and Ofek, Eran and Szpektor, Idan},
  journal={Advances in Neural Information Processing Systems},
  volume={36},
  pages={1601--1619},
  year={2023}
}

@article{dsg,
  title={Davidsonian scene graph: Improving reliability in fine-grained evaluation for text-to-image generation},
  author={Cho, Jaemin and Hu, Yushi and Garg, Roopal and Anderson, Peter and Krishna, Ranjay and Baldridge, Jason and Bansal, Mohit and Pont-Tuset, Jordi and Wang, Su},
  journal={arXiv preprint arXiv:2310.18235},
  year={2023}
}

@article{vpeval,
  title={Visual programming for step-by-step text-to-image generation and evaluation},
  author={Cho, Jaemin and Zala, Abhay and Bansal, Mohit},
  journal={Advances in Neural Information Processing Systems},
  volume={36},
  pages={6048--6069},
  year={2023}
}

@inproceedings{hallu,
  title={Evaluating Image Hallucination in Text-to-Image Generation with Question-Answering},
  author={Lim, Youngsun and Choi, Hojun and Shim, Hyunjung},
  booktitle={Proceedings of the AAAI Conference on Artificial Intelligence},
  volume={39},
  number={25},
  pages={26290--26298},
  year={2025}
}

@article{chen2024mllmjudge,
  title={MLLM-as-a-Judge: Assessing Multimodal LLM-as-a-Judge with Vision-Language Benchmark},
  author={Chen, Dongping and others},
  journal={arXiv preprint arXiv:2402.04788},
  year={2024}
}

@article{zhang2025abench,
  title={A-Bench: Are LMMs Masters at Evaluating AI-generated Images?},
  author={Zhang, Zicheng and others},
  journal={arXiv preprint arXiv:2406.03070},
  year={2025}
}

@inproceedings{chen2025mjbench,
  title={MJ-Bench: Is Your Multimodal Reward Model Really a Good Judge for Text-to-Image Generation?},
  author={Chen, Zhaorun and others},
  booktitle={Advances in Neural Information Processing Systems},
  year={2025}
}

@article{tong2025mjvideo,
  title={MJ-VIDEO: Fine-Grained Benchmarking and Rewarding Video Preferences in Video Generation},
  author={Tong, Haibo and others},
  journal={arXiv preprint arXiv:2502.01719},
  year={2025}
}

@inproceedings{sd3medium,
  title={Scaling rectified flow transformers for high-resolution image synthesis},
  author={Esser, Patrick and Kulal, Sumith and Blattmann, Andreas and Entezari, Rahim and M{\"u}ller, Jonas and Saini, Harry and Levi, Yam and Lorenz, Dominik and Sauer, Axel and Boesel, Frederic and others},
  booktitle={Forty-first International Conference on Machine Learning},
  year={2024}
}

@article{playground,
  title={Playground v2. 5: Three insights towards enhancing aesthetic quality in text-to-image generation},
  author={Li, Daiqing and Kamko, Aleks and Akhgari, Ehsan and Sabet, Ali and Xu, Linmiao and Doshi, Suhail},
  journal={arXiv preprint arXiv:2402.17245},
  year={2024}
}

@misc{flux2024,
    author={Black Forest Labs},
    title={FLUX},
    year={2024},
    howpublished={\url{https://github.com/black-forest-labs/flux}},
}

@misc{hidream,
    author={hidream},
    title={hidream},
    year={2024},
    howpublished={\url{https://github.com/HiDream-ai/HiDream-I1}},
}

@article{janus,
  title={Janus-Pro: Unified Multimodal Understanding and Generation with Data and Model Scaling},
  author={Chen, Xiaokang and Wu, Zhiyu and Liu, Xingchao and Pan, Zizheng and Liu, Wen and Xie, Zhenda and Yu, Xingkai and Ruan, Chong},
  journal={arXiv preprint arXiv:2501.17811},
  year={2025}
}

@article{showo,
  title={Show-o: One single transformer to unify multimodal understanding and generation},
  author={Xie, Jinheng and Mao, Weijia and Bai, Zechen and Zhang, David Junhao and Wang, Weihao and Lin, Kevin Qinghong and Gu, Yuchao and Chen, Zhijie and Yang, Zhenheng and Shou, Mike Zheng},
  journal={arXiv preprint arXiv:2408.12528},
  year={2024}
}

@article{chen2025blip3,
  title={Blip3-o: A family of fully open unified multimodal models-architecture, training and dataset},
  author={Chen, Jiuhai and Xu, Zhiyang and Pan, Xichen and Hu, Yushi and Qin, Can and Goldstein, Tom and Huang, Lifu and Zhou, Tianyi and Xie, Saining and Savarese, Silvio and others},
  journal={arXiv preprint arXiv:2505.09568},
  year={2025}
}

@article{emu3,
  title={Emu3: Next-token prediction is all you need},
  author={Wang, Xinlong and Zhang, Xiaosong and Luo, Zhengxiong and Sun, Quan and Cui, Yufeng and Wang, Jinsheng and Zhang, Fan and Wang, Yueze and Li, Zhen and Yu, Qiying and others},
  journal={arXiv preprint arXiv:2409.18869},
  year={2024}
}

@article{got,
  title={Got: Unleashing reasoning capability of multimodal large language model for visual generation and editing},
  author={Fang, Rongyao and Duan, Chengqi and Wang, Kun and Huang, Linjiang and Li, Hao and Yan, Shilin and Tian, Hao and Zeng, Xingyu and Zhao, Rui and Dai, Jifeng and others},
  journal={arXiv preprint arXiv:2503.10639},
  year={2025}
}

@article{duan2025got,
  title={Got-r1: Unleashing reasoning capability of mllm for visual generation with reinforcement learning},
  author={Duan, Chengqi and Fang, Rongyao and Wang, Yuqing and Wang, Kun and Huang, Linjiang and Zeng, Xingyu and Li, Hongsheng and Liu, Xihui},
  journal={arXiv preprint arXiv:2505.17022},
  year={2025}
}

@misc{GPT,
    author  = {OpenAI},
    title = {GPT-Image-1},
    year = {2025},
    howpublished = {\url{https://openai.com/index/image-generation-api/}}
    
}

@misc{nano-banana,
  title={Introducing Gemini 2.5 Flash Image, our state-of-the-art image model},
  author={Google},
  year={2025},
  howpublished ={\url{https://developers.googleblog.com/en/introducing-gemini-2-5-flash-image/}}
}

@article{seedream2025,
  title={Seedream 4.0: Toward Next-generation Multimodal Image Generation},
  author={{Team Seedream}},
  journal={arXiv preprint arXiv:2509.20427},
  year={2025},
  eprint={2509.20427},
  archivePrefix={arXiv},
  primaryClass={cs.CV},
  url={https://arxiv.org/abs/2509.20427}
}

@article{wu2025qwen,
  title={Qwen-image technical report},
  author={Wu, Chenfei and Li, Jiahao and Zhou, Jingren and Lin, Junyang and Gao, Kaiyuan and Yan, Kun and Yin, Sheng-ming and Bai, Shuai and Xu, Xiao and Chen, Yilei and others},
  journal={arXiv preprint arXiv:2508.02324},
  year={2025}
}

@article{deng2025emerging,
  title={Emerging properties in unified multimodal pretraining},
  author={Deng, Chaorui and Zhu, Deyao and Li, Kunchang and Gou, Chenhui and Li, Feng and Wang, Zeyu and Zhong, Shu and Yu, Weihao and Nie, Xiaonan and Song, Ziang and others},
  journal={arXiv preprint arXiv:2505.14683},
  year={2025}
}

@misc{idiombook,
    author  = {},
    title = {The Exhaustive List of American Idioms},
    year = {2023},
    howpublished = {\url{https://learn-esl.org/index.html}}
    
}

@article{Qwen2.5-VL,
  title={Qwen2.5-VL Technical Report},
  author={Bai, Shuai and Chen, Keqin and Liu, Xuejing and Wang, Jialin and Ge, Wenbin and Song, Sibo and Dang, Kai and Wang, Peng and Wang, Shijie and Tang, Jun and Zhong, Humen and Zhu, Yuanzhi and Yang, Mingkun and Li, Zhaohai and Wan, Jianqiang and Wang, Pengfei and Ding, Wei and Fu, Zheren and Xu, Yiheng and Ye, Jiabo and Zhang, Xi and Xie, Tianbao and Cheng, Zesen and Zhang, Hang and Yang, Zhibo and Xu, Haiyang and Lin, Junyang},
  journal={arXiv preprint arXiv:2502.13923},
  year={2025}
}

@article{guo2025deepseek,
  title={Deepseek-r1: Incentivizing reasoning capability in llms via reinforcement learning},
  author={Guo, Daya and Yang, Dejian and Zhang, Haowei and Song, Junxiao and Zhang, Ruoyu and Xu, Runxin and Zhu, Qihao and Ma, Shirong and Wang, Peiyi and Bi, Xiao and others},
  journal={arXiv preprint arXiv:2501.12948},
  year={2025}
}

@inproceedings{wei2022chain,
  title={Chain-of-thought prompting elicits reasoning in large language models},
  author={Wei, Jason and Wang, Xuezhi and Schuurmans, Dale and Bosma, Maarten and Xia, Fei and Chi, Ed and Le, Quoc V and Zhou, Denny and others},
  booktitle={NeurIPS},
  year={2022}
}

@article{zhou2024high,
  title={A High-Quality Text-Rich Image Instruction Tuning Dataset via Hybrid Instruction Generation},
  author={Zhou, Shijie and Zhang, Ruiyi and Zhou, Yufan and Chen, Changyou},
  journal={arXiv preprint arXiv:2412.16364},
  year={2024}
}

@article{schuhmann2022laion,
  title={Laion-5b: An open large-scale dataset for training next generation image-text models},
  author={Schuhmann, Christoph and Beaumont, Romain and Vencu, Richard and Gordon, Cade and Wightman, Ross and Cherti, Mehdi and Coombes, Theo and Katta, Aarush and Mullis, Clayton and Wortsman, Mitchell and others},
  journal={Advances in neural information processing systems},
  volume={35},
  pages={25278--25294},
  year={2022}
}

@inproceedings{mathew2022infographicvqa,
  title={Infographicvqa},
  author={Mathew, Minesh and Bagal, Viraj and Tito, Rub{\`e}n and Karatzas, Dimosthenis and Valveny, Ernest and Jawahar, CV},
  booktitle={Proceedings of the IEEE/CVF Winter Conference on Applications of Computer Vision},
  pages={1697--1706},
  year={2022}
}

@article{chen2025posta,
  title={Posta: A go-to framework for customized artistic poster generation},
  author={Chen, Haoyu and Xu, Xiaojie and Li, Wenbo and Ren, Jingjing and Ye, Tian and Liu, Songhua and Chen, Ying-Cong and Zhu, Lei and Wang, Xinchao},
  journal={arXiv preprint arXiv:2503.14908},
  year={2025}
}

@article{yang2025scaling,
  title={Scaling text-rich image understanding via code-guided synthetic multimodal data generation},
  author={Yang, Yue and Patel, Ajay and Deitke, Matt and Gupta, Tanmay and Weihs, Luca and Head, Andrew and Yatskar, Mark and Callison-Burch, Chris and Krishna, Ranjay and Kembhavi, Aniruddha and others},
  journal={arXiv preprint arXiv:2502.14846},
  year={2025}
}

@article{gpt4o,
  title={Gpt-4o system card},
  author={Hurst, Aaron and Lerer, Adam and Goucher, Adam P and Perelman, Adam and Ramesh, Aditya and Clark, Aidan and Ostrow, AJ and Welihinda, Akila and Hayes, Alan and Radford, Alec and others},
  journal={arXiv preprint arXiv:2410.21276},
  year={2024}
}

@article{hessel2021clipscore,
  title={Clipscore: A reference-free evaluation metric for image captioning},
  author={Hessel, Jack and Holtzman, Ari and Forbes, Maxwell and Bras, Ronan Le and Choi, Yejin},
  journal={arXiv preprint arXiv:2104.08718},
  year={2021}
}

@inproceedings{lin2024evaluating,
  title={Evaluating text-to-visual generation with image-to-text generation},
  author={Lin, Zhiqiu and Pathak, Deepak and Li, Baiqi and Li, Jiayao and Xia, Xide and Neubig, Graham and Zhang, Pengchuan and Ramanan, Deva},
  booktitle={European Conference on Computer Vision},
  pages={366--384},
  year={2024},
  organization={Springer}
}

@misc{GPT-5.1,
  title={GPT-5.1: A smarter, more conversational ChatGPT},
  author={OpenAI},
  year={2025},
  howpublished ={\url{https://openai.com/index/gpt-5-1/}}
}

@misc{Gemini-2.5-pro,
  title={{Gemini 2.5}: Our most intelligent {AI} model},
  author={{Google DeepMind}},
  year={2025},
  howpublished={\url{https://blog.google/technology/google-deepmind/gemini-model-thinking-updates-march-2025/}}
}

@article{bai2025qwen3,
  title={Qwen3-VL Technical Report},
  author={Bai, Shuai and Cai, Yuxuan and Chen, Ruizhe and Chen, Keqin and Chen, Xionghui and Cheng, Zesen and Deng, Lianghao and Ding, Wei and Gao, Chang and Ge, Chunjiang and others},
  journal={arXiv preprint arXiv:2511.21631},
  year={2025}
}

@article{an2025llava,
  title={Llava-onevision-1.5: Fully open framework for democratized multimodal training},
  author={An, Xiang and Xie, Yin and Yang, Kaicheng and Zhang, Wenkang and Zhao, Xiuwei and Cheng, Zheng and Wang, Yirui and Xu, Songcen and Chen, Changrui and Wu, Chunsheng and others},
  journal={arXiv preprint arXiv:2509.23661},
  year={2025}
}

\clearpage
\appendix

\section*{Appendix Contents}
\startcontents
\printcontents{ }{1}{}

\section{Detailed Component Scores}
\label{app:component_scores}
Tables~\ref{tab:components_original} and~\ref{tab:components_pipeline} report all component scores for the 16 models under original and LLM-rewritten prompts, respectively. Each component is averaged across questions within an image and then across the evaluated images. Category totals use $0.9R+0.1Q$ for Idiom Interpretation, $0.7R+0.2T+0.1Q$ for Textual Image Design, and $0.7R+0.2D+0.1Q$ for Entity and Scientific Reasoning, where $R$, $T$, $D$, and $Q$ denote reasoning, text rendering, other details, and image quality.

\noindent\textbf{Prompt rewriting primarily improves reasoning rather than visual quality.}
With original prompts, Qwen-Image scores 95.3 and 92.6 on image quality for Idiom Interpretation and Scientific Reasoning, but only 59.0 and 61.3 on their reasoning components. Rewriting raises these reasoning scores to 83.4 and 77.5, and entity reasoning from 51.8 to 80.8. The corresponding quality changes are smaller (95.3 to 96.2, 92.6 to 97.0, and 93.7 to 92.2), indicating that these gains primarily concern semantic correctness rather than visual polish.

\noindent\textbf{Separating text rendering from design reasoning.}
Text rendering and design reasoning exhibit distinct performance patterns. With original prompts, SD-3.5-Large achieves a higher design reasoning score than FLUX.1-dev (74.3 vs. 71.4), but substantially lower text rendering performance (39.0 vs. 62.1). Similarly, Qwen-Image scores 81.2 on design reasoning but only 58.0 on text rendering. These contrasting profiles suggest that the ability to fulfill a design intention is distinct from the ability to render readable, correct text. Evaluating text rendering separately therefore helps distinguish rendering limitations from reasoning failures, providing a clearer assessment of design reasoning.
\begin{table*}[t]
\centering
\caption{\textbf{Component scores with T2I-ReasonBench.} All scores are scaled to 0--100. R: reasoning; T: text rendering; D: other details; Q: image quality. \colorbox{softgray}{Shaded} columns report the weighted category totals. \textbf{Bold} marks the highest mean in each column.}
\label{tab:components_original}
\begingroup
\setlength{\tabcolsep}{2.2pt}
\setlength{\fboxsep}{2pt}
\renewcommand{\arraystretch}{1.16}
\footnotesize
\resizebox{\textwidth}{!}{%
\begin{tabular}{lccc|cccc|cccc|cccc}
\toprule
\multirow{2}{*}{\textbf{Model}} & \multicolumn{3}{c}{\textbf{Idiom Interpretation}} & \multicolumn{4}{c}{\textbf{Textual Image Design}} & \multicolumn{4}{c}{\textbf{Entity Reasoning}} & \multicolumn{4}{c}{\textbf{Scientific Reasoning}} \\
\cmidrule(lr){2-4}\cmidrule(lr){5-8}\cmidrule(lr){9-12}\cmidrule(lr){13-16}
 & R & Q & \colorbox{softgray}{Overall} & R & T & Q & \colorbox{softgray}{Overall} & R & D & Q & \colorbox{softgray}{Overall} & R & D & Q & \colorbox{softgray}{Overall} \\
\midrule
\multicolumn{16}{c}{\textit{Diffusion Models}} \\
\midrule
SD-3-Medium & 37.7 & 77.9 & \colorbox{softgray}{41.7} & 72.5 & 48.7 & 87.9 & \colorbox{softgray}{69.3} & 30.7 & 64.8 & 86.4 & \colorbox{softgray}{43.1} & 44.1 & 71.6 & 86.4 & \colorbox{softgray}{53.8} \\
SD-3.5-Medium & 35.1 & 80.0 & \colorbox{softgray}{39.6} & 71.4 & 45.7 & 92.4 & \colorbox{softgray}{68.4} & 35.6 & 68.3 & 90.1 & \colorbox{softgray}{47.6} & 44.7 & 70.5 & 87.9 & \colorbox{softgray}{54.1} \\
SD-3.5-Large & 39.3 & 88.7 & \colorbox{softgray}{44.2} & 74.3 & 39.0 & 90.0 & \colorbox{softgray}{68.8} & 39.2 & 70.2 & 94.7 & \colorbox{softgray}{50.9} & 49.5 & 75.5 & 88.3 & \colorbox{softgray}{58.6} \\
FLUX.1-dev & 45.2 & 86.6 & \colorbox{softgray}{49.3} & 71.4 & 62.1 & 96.8 & \colorbox{softgray}{72.1} & 40.8 & 64.8 & 90.0 & \colorbox{softgray}{50.5} & 46.7 & 74.8 & 86.8 & \colorbox{softgray}{56.3} \\
FLUX.1-schnell & 43.4 & 83.8 & \colorbox{softgray}{47.4} & 79.5 & 66.1 & 95.0 & \colorbox{softgray}{78.3} & 39.8 & 67.6 & 87.2 & \colorbox{softgray}{50.1} & 52.3 & 78.7 & 87.3 & \colorbox{softgray}{61.1} \\
Playground-v2.5 & 46.4 & 90.8 & \colorbox{softgray}{50.8} & 52.1 & 27.3 & 92.8 & \colorbox{softgray}{51.2} & 40.3 & 70.3 & 87.7 & \colorbox{softgray}{51.1} & 42.5 & 63.4 & 82.6 & \colorbox{softgray}{50.7} \\
HiDream-I1-full & 55.9 & 87.9 & \colorbox{softgray}{59.1} & 81.9 & 73.6 & 97.4 & \colorbox{softgray}{81.8} & 48.4 & 76.2 & 89.0 & \colorbox{softgray}{58.0} & 51.2 & 75.6 & 89.4 & \colorbox{softgray}{59.9} \\
Qwen-Image & 59.0 & 95.3 & \colorbox{softgray}{62.7} & 81.2 & 58.0 & 91.9 & \colorbox{softgray}{77.6} & 51.8 & 81.0 & 93.7 & \colorbox{softgray}{61.8} & 61.3 & 84.1 & 92.6 & \colorbox{softgray}{69.0} \\
\midrule
\multicolumn{16}{c}{\textit{Unified Multimodal Models}} \\
\midrule
Emu3 & 34.8 & 79.4 & \colorbox{softgray}{39.2} & 46.1 & 22.9 & 78.1 & \colorbox{softgray}{44.6} & 32.4 & 61.1 & 77.9 & \colorbox{softgray}{42.7} & 34.6 & 56.5 & 73.3 & \colorbox{softgray}{42.9} \\
show-o-demo-512 & 34.5 & 78.7 & \colorbox{softgray}{38.9} & 46.4 & 26.6 & 93.3 & \colorbox{softgray}{47.2} & 30.7 & 56.8 & 81.5 & \colorbox{softgray}{41.0} & 39.4 & 66.6 & 83.7 & \colorbox{softgray}{49.2} \\
Janus-Pro-7B & 28.3 & 72.3 & \colorbox{softgray}{32.7} & 46.7 & 25.7 & 79.2 & \colorbox{softgray}{45.8} & 30.8 & 61.0 & 75.3 & \colorbox{softgray}{41.3} & 37.4 & 57.5 & 67.3 & \colorbox{softgray}{44.4} \\
GoT & 28.6 & 79.4 & \colorbox{softgray}{33.7} & 43.9 & 45.3 & 91.1 & \colorbox{softgray}{48.9} & 29.0 & 46.8 & 78.8 & \colorbox{softgray}{37.6} & 34.3 & 54.2 & 79.8 & \colorbox{softgray}{42.8} \\
Bagel w/ Thinking & 46.3 & 89.2 & \colorbox{softgray}{50.6} & 54.7 & 45.6 & 94.2 & \colorbox{softgray}{56.8} & 49.4 & 66.0 & 86.6 & \colorbox{softgray}{56.5} & 58.9 & 80.0 & 92.8 & \colorbox{softgray}{66.5} \\
\midrule
\multicolumn{16}{c}{\textit{Proprietary Models}} \\
\midrule
Seedream 4.0 & 83.3 & 97.4 & \colorbox{softgray}{84.8} & \textbf{97.0} & 84.4 & 92.8 & \colorbox{softgray}{94.1} & 66.8 & 83.4 & 95.2 & \colorbox{softgray}{73.0} & 80.8 & 90.3 & 97.3 & \colorbox{softgray}{84.4} \\
GPT-Image-1 & 82.5 & \textbf{98.1} & \colorbox{softgray}{84.1} & 93.5 & \textbf{95.7} & \textbf{98.2} & \colorbox{softgray}{\textbf{94.4}} & 82.7 & 90.5 & \textbf{97.5} & \colorbox{softgray}{85.7} & 83.7 & 93.5 & 97.7 & \colorbox{softgray}{87.1} \\
Nano-Banana & \textbf{89.0} & 96.9 & \colorbox{softgray}{\textbf{89.8}} & 95.8 & 86.6 & 96.6 & \colorbox{softgray}{94.0} & \textbf{83.8} & \textbf{95.0} & 96.4 & \colorbox{softgray}{\textbf{87.3}} & \textbf{87.4} & \textbf{94.4} & \textbf{98.5} & \colorbox{softgray}{\textbf{89.9}} \\
\bottomrule
\end{tabular}%
}
\endgroup
\end{table*}

\begin{table*}[!t]
\centering
\caption{\textbf{Component scores with LLM-rewritten prompts.} All scores are scaled to 0--100. R: reasoning; T: text rendering; D: other details; Q: image quality. \colorbox{softgray}{Shaded} columns report the weighted category totals. \textbf{Bold} marks the highest mean in each column.}
\label{tab:components_pipeline}
\begingroup
\setlength{\tabcolsep}{2.2pt}
\setlength{\fboxsep}{2pt}
\renewcommand{\arraystretch}{1.16}
\footnotesize
\resizebox{\textwidth}{!}{%
\begin{tabular}{lccc|cccc|cccc|cccc}
\toprule
\multirow{2}{*}{\textbf{Model}} & \multicolumn{3}{c}{\textbf{Idiom Interpretation}} & \multicolumn{4}{c}{\textbf{Textual Image Design}} & \multicolumn{4}{c}{\textbf{Entity Reasoning}} & \multicolumn{4}{c}{\textbf{Scientific Reasoning}} \\
\cmidrule(lr){2-4}\cmidrule(lr){5-8}\cmidrule(lr){9-12}\cmidrule(lr){13-16}
 & R & Q & \colorbox{softgray}{Overall} & R & T & Q & \colorbox{softgray}{Overall} & R & D & Q & \colorbox{softgray}{Overall} & R & D & Q & \colorbox{softgray}{Overall} \\
\midrule
\multicolumn{16}{c}{\textit{Diffusion Models}} \\
\midrule
SD-3-Medium & 75.5 & 87.6 & \colorbox{softgray}{76.7} & 79.7 & 54.9 & 92.6 & \colorbox{softgray}{76.0} & 71.8 & 82.0 & 84.0 & \colorbox{softgray}{75.0} & 62.4 & 77.9 & 94.2 & \colorbox{softgray}{68.7} \\
SD-3.5-Medium & 74.8 & 91.1 & \colorbox{softgray}{76.4} & 78.7 & 55.4 & 93.9 & \colorbox{softgray}{75.6} & 72.5 & 82.6 & 90.9 & \colorbox{softgray}{76.4} & 62.4 & 77.4 & 95.1 & \colorbox{softgray}{68.7} \\
SD-3.5-Large & 74.8 & 94.1 & \colorbox{softgray}{76.7} & 80.6 & 49.7 & 94.2 & \colorbox{softgray}{75.8} & 79.5 & 85.6 & 93.2 & \colorbox{softgray}{82.1} & 67.3 & 79.8 & 95.5 & \colorbox{softgray}{72.6} \\
FLUX.1-dev & 76.0 & 94.1 & \colorbox{softgray}{77.8} & 80.7 & 70.5 & 97.0 & \colorbox{softgray}{80.3} & 71.4 & 83.8 & 88.6 & \colorbox{softgray}{75.6} & 69.1 & 80.8 & 94.8 & \colorbox{softgray}{74.0} \\
FLUX.1-schnell & 78.7 & 90.2 & \colorbox{softgray}{79.9} & 82.0 & 67.9 & 95.2 & \colorbox{softgray}{80.5} & 72.9 & 85.2 & 85.0 & \colorbox{softgray}{76.6} & 69.8 & 82.3 & 95.7 & \colorbox{softgray}{74.9} \\
Playground-v2.5 & 62.5 & 90.2 & \colorbox{softgray}{65.3} & 53.9 & 32.0 & 93.6 & \colorbox{softgray}{53.5} & 76.3 & 76.7 & 86.9 & \colorbox{softgray}{77.4} & 49.1 & 63.3 & 85.2 & \colorbox{softgray}{55.6} \\
HiDream-I1-full & 74.9 & 95.6 & \colorbox{softgray}{77.0} & 85.2 & 76.2 & 97.0 & \colorbox{softgray}{84.6} & 77.1 & 84.4 & 88.0 & \colorbox{softgray}{79.6} & 65.2 & 79.3 & 95.4 & \colorbox{softgray}{71.0} \\
Qwen-Image & 83.4 & 96.2 & \colorbox{softgray}{84.7} & 86.2 & 64.1 & 94.9 & \colorbox{softgray}{82.7} & 80.8 & 92.2 & 92.2 & \colorbox{softgray}{84.2} & 77.5 & 87.3 & 97.0 & \colorbox{softgray}{81.4} \\
\midrule
\multicolumn{16}{c}{\textit{Unified Multimodal Models}} \\
\midrule
Emu3 & 66.2 & 81.4 & \colorbox{softgray}{67.7} & 55.1 & 26.9 & 83.6 & \colorbox{softgray}{52.4} & 65.1 & 73.5 & 73.7 & \colorbox{softgray}{67.6} & 48.0 & 62.0 & 81.6 & \colorbox{softgray}{54.2} \\
show-o-demo-512 & 74.1 & 81.2 & \colorbox{softgray}{74.8} & 53.1 & 22.3 & 95.3 & \colorbox{softgray}{51.2} & 66.0 & 77.5 & 76.3 & \colorbox{softgray}{69.4} & 57.9 & 72.7 & 93.0 & \colorbox{softgray}{64.4} \\
Janus-Pro-7B & 72.4 & 75.0 & \colorbox{softgray}{72.6} & 62.7 & 23.4 & 88.4 & \colorbox{softgray}{57.4} & 68.0 & 80.8 & 72.5 & \colorbox{softgray}{71.0} & 60.0 & 72.2 & 84.9 & \colorbox{softgray}{64.9} \\
GoT & 59.8 & 85.4 & \colorbox{softgray}{62.3} & 48.6 & 46.6 & 93.7 & \colorbox{softgray}{52.7} & 48.4 & 66.7 & 79.8 & \colorbox{softgray}{55.2} & 39.5 & 57.6 & 84.6 & \colorbox{softgray}{47.6} \\
Bagel w/o Thinking & 76.5 & 90.6 & \colorbox{softgray}{77.9} & 72.0 & 56.6 & 96.5 & \colorbox{softgray}{71.4} & 70.0 & 82.9 & 83.0 & \colorbox{softgray}{73.9} & 70.6 & 80.6 & 95.6 & \colorbox{softgray}{75.1} \\
\midrule
\multicolumn{16}{c}{\textit{Proprietary Models}} \\
\midrule
Seedream 4.0 & 86.1 & 98.0 & \colorbox{softgray}{87.3} & 89.6 & 75.4 & 93.2 & \colorbox{softgray}{87.1} & 84.0 & 92.5 & 95.2 & \colorbox{softgray}{86.8} & 85.9 & 89.3 & 99.0 & \colorbox{softgray}{87.9} \\
GPT-Image-1 & 85.9 & \textbf{98.4} & \colorbox{softgray}{87.1} & 90.1 & \textbf{96.7} & \textbf{98.4} & \colorbox{softgray}{92.3} & \textbf{89.2} & \textbf{94.1} & 95.1 & \colorbox{softgray}{\textbf{90.8}} & \textbf{87.3} & \textbf{91.4} & 99.0 & \colorbox{softgray}{\textbf{89.3}} \\
Nano-Banana & \textbf{86.6} & 97.3 & \colorbox{softgray}{\textbf{87.7}} & \textbf{93.0} & 89.9 & 96.5 & \colorbox{softgray}{\textbf{92.7}} & 88.5 & 92.2 & \textbf{98.0} & \colorbox{softgray}{90.2} & 87.1 & 90.6 & \textbf{99.4} & \colorbox{softgray}{89.1} \\
\bottomrule
\end{tabular}%
}
\endgroup
\end{table*}

\begin{figure*}[!t]
    \includegraphics[width=\linewidth]{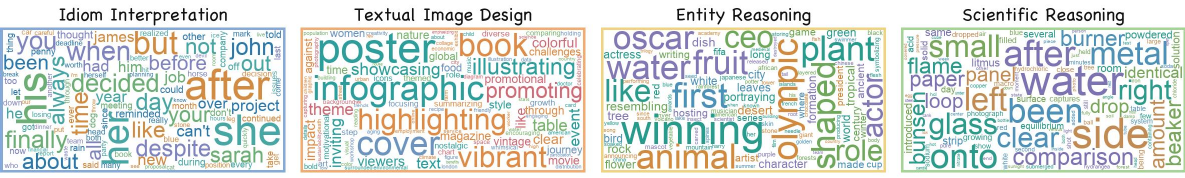}
    \caption{Word cloud to visualize the word distribution of each dimension in our prompt suite.}
    \label{fig:word_cloud}
\end{figure*}

\section{Qualitative Examples}
\label{app:qualitative}
\begin{figure*}[h!]
    \centering
    \includegraphics[width=\linewidth]{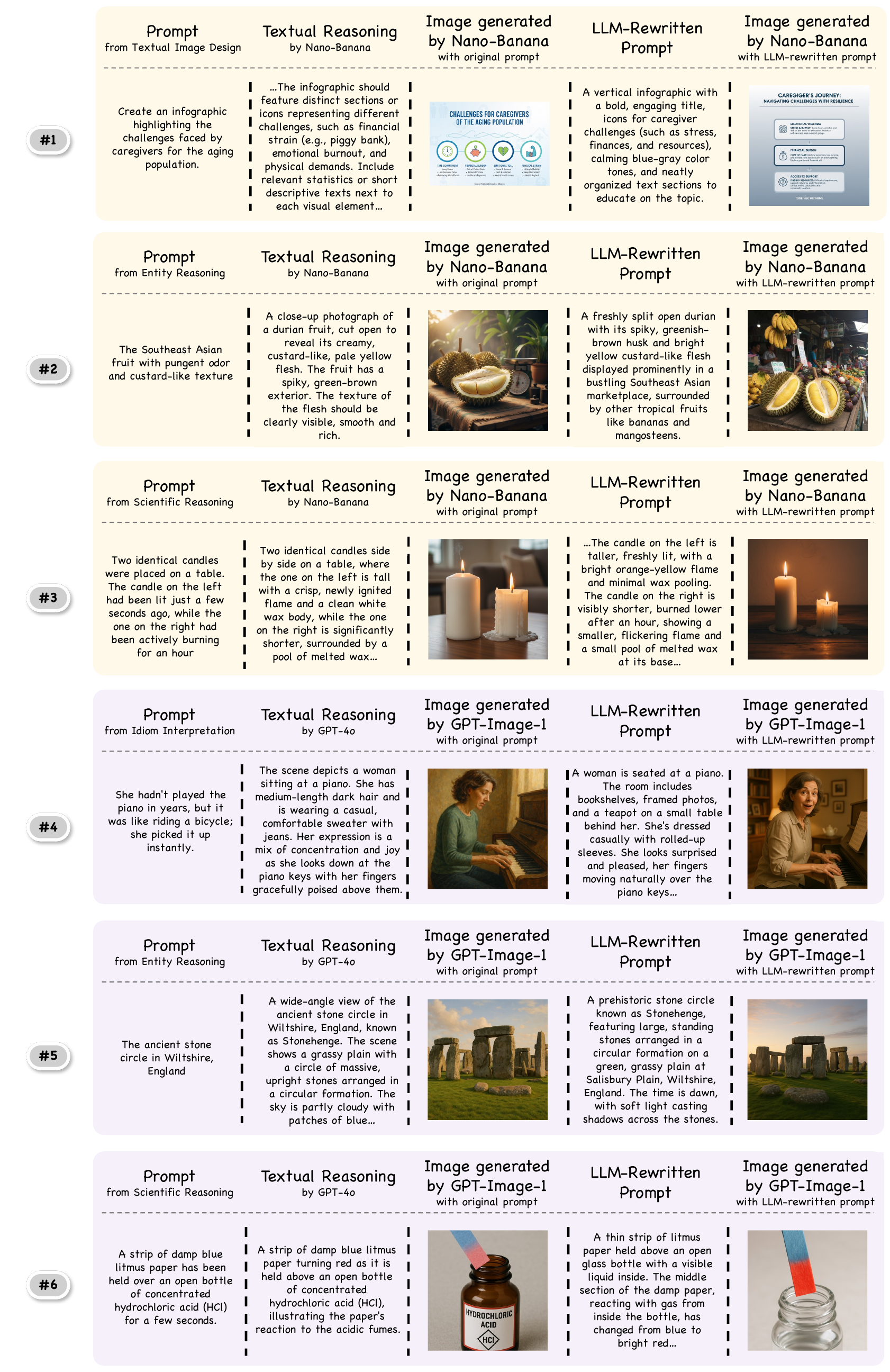}
    \caption{Qualitative examples with original prompt and generated image, textual reasoning by text module, LLM-rewritten prompt and generated image.}
    \vspace{-10pt}
    \label{fig:good_quality}
\end{figure*}

\begin{figure*}[h!]
     \centering
    \includegraphics[width=\linewidth]{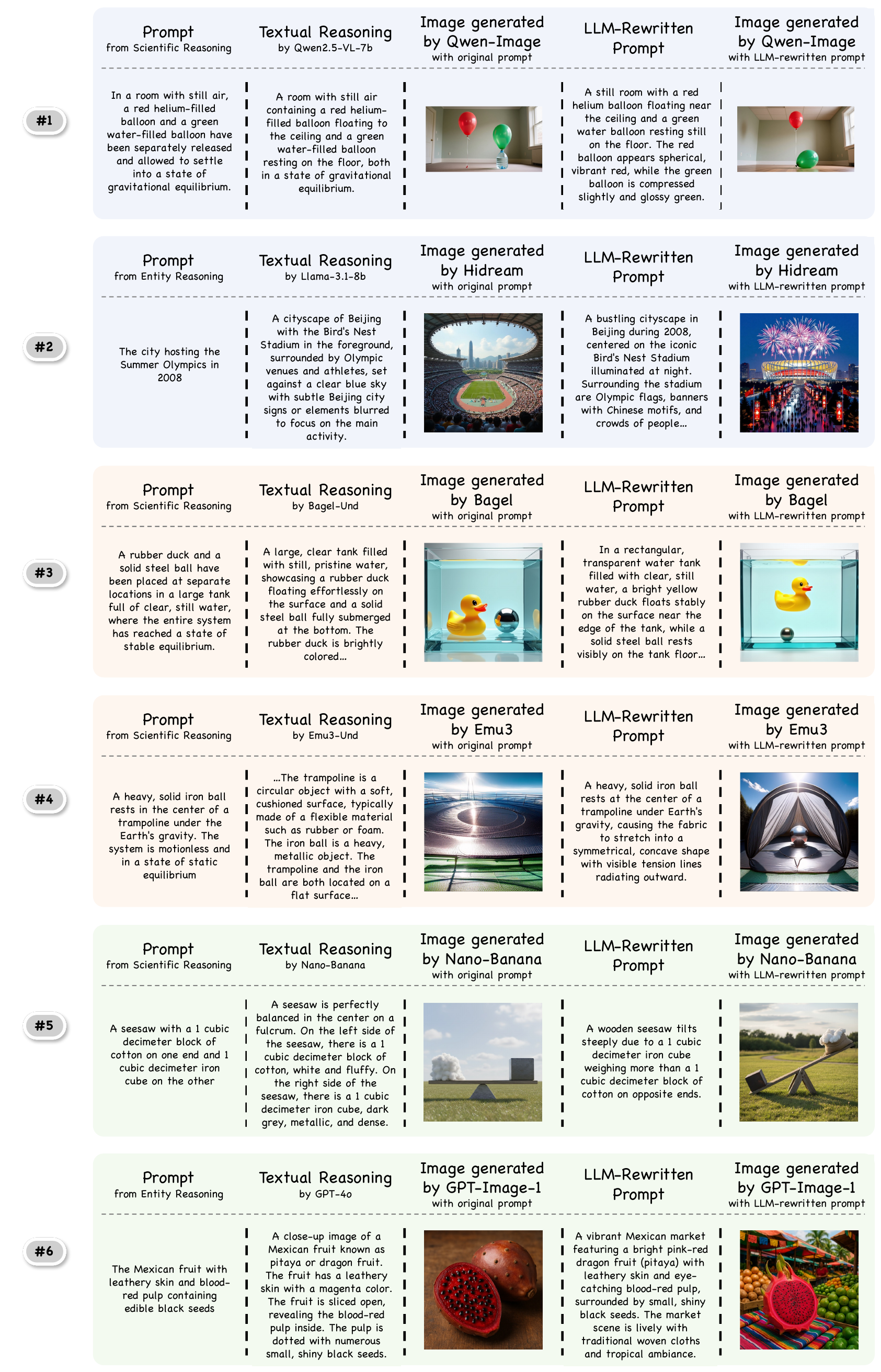}
    \caption{Qualitative examples with original prompt and generated image, textual reasoning by text module, LLM-rewritten prompt and generated image.}
    \vspace{-10pt}
    \label{fig:bad_quality}
\end{figure*}

\begin{figure*}[h!]
     \centering
    \includegraphics[width=\linewidth]{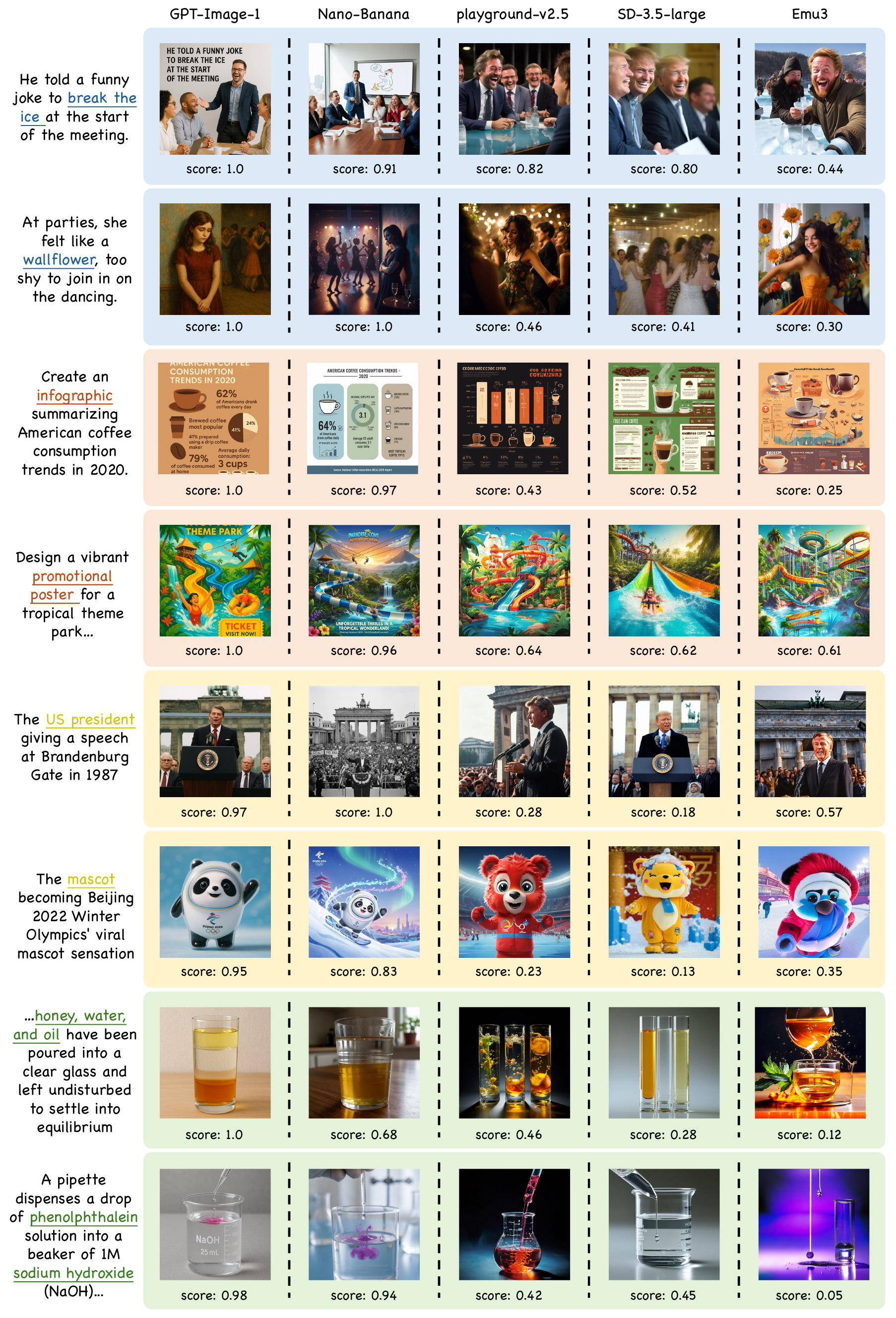}
    \caption{Qualitative examples with evaluated T2I-ReasonScore.}
    \vspace{-10pt}
    \label{fig:qualitative}
\end{figure*}

Figure~\ref{fig:good_quality} and ~\ref{fig:bad_quality} compare the qualitative outputs of different T2I models.
Figure~\ref{fig:qualitative} shows more image outputs from different models with the evaluated T2I-ReasonScore.

\section{More Details on Prompt Collection Process}
\label{app:prompt}
\textbf{Idiom Interpretation.} 
In idiom collection, we leverage a book titled ``The Exhaustive List of
American Idioms''~\cite{idiombook}, which systematically documents over 11k idioms. These idioms were collected from diverse sources, including TV shows, movies, and everyday conversations. Each idiom in the book is accompanied by its actual meaning in context. In addition to this resource, we also refer to idioms available on the Internet. From this extensive pool, we manually select 200 idioms that are commonly used in daily life and challenging for T2I models due to their figurative meanings. We input the selected idioms and their actual meanings into an LLM and prompt it to generate new sentences. These sentences are designed to describe visible scenes involving the idioms, providing contextual clues for reasoning while avoiding directly revealing the idiom's meaning.

\noindent \textbf{Textual Image Design.}
For textual image design, we collect 6 types of text-rich images from 4
distinct sources. 

(1) LLAVAR-2 Dataset~\cite{zhou2024high}: This dataset contains 42k text-rich images sourced from LAION~\cite{schuhmann2022laion}, representing various categories such as quotes, memes, book covers, posters, and product packaging. However, images in this dataset are of various quality and formats, so we filter out 80 high-quality images that have a resolution greater than 384x384 and exhibit clear design intentions.

(2) InfographicVQA Dataset~\cite{mathew2022infographicvqa}: This dataset comprises 5k high-quality infographics. We select 40 with normal height-width ratio that exemplify well-crafted layouts to convey structured information.

(3) POSTA Dataset~\cite{chen2025posta}: This dataset includes over 300 posters with professional background, layout, and text formats designed by experts. We select 40 posters that demonstrate a balance between text and visual design elements.

(4) CoSyn-400k Dataset~\cite{yang2025scaling}: This dataset consists of 400k synthetic text-rich images, generated by LLM-drive codes.
These images cover diverse formats, such as charts, diagrams, tables, documents (e.g., menus or business cards), math examples, and musical scores. From this dataset, we select 40 samples that exemplify structured text-visual integration, including 10 tables, 10 diagrams, and 20 documents.

The design intentions of the collected images are summarized using Qwen2.5-VL~\cite{Qwen2.5-VL}, yielding 200 prompts in this dimension.

\noindent \textbf{Entity Reasoning and Scientific Reasoning.} 
Our prompt generation pipeline for each of these two dimensions consists of three stages: initial design, LLM-based expansion, and expert verification.
The process begins with human experts defining key subcategories (e.g., different scientific disciplines, or well-known entities with clear visual features). For each subcategory, the experts then authored a set of high-quality ``seed'' prompts to serve as a foundation.
Next, these seed prompts were used as exemplars to prompt multiple LLMs to generate a larger, more diverse corpus of prompts. The LLMs also generated an explicit description of the expected visual phenomena to assist with evaluation.
Finally, all generated prompts and their explicit descriptions underwent a rigorous verification by human experts. This review includes consulting reliable sources to fact-check all domain-specific information and ensure its accuracy. \\
We visualize the word distribution of each dimension in our prompt suite in Figure~\ref{fig:word_cloud}.

\section{Evaluation Framework}
\label{app:evaluation}
\subsection{Templates}
\textbf{Templates for evaluation stage I:  LLM question-criterion generation.}
Table~\ref{tab:qc_1}, ~\ref{tab:qc_2}, ~\ref{tab:qc_3}, and~\ref{tab:qc_4} present the templates used to generate the prompt-specific question-criterion pairs for Idiom Interpretation, Textual Image Design, Entity-Reasoning and Scientific-Reasoning, respectively. Each template is tailored to focus on the unique aspects of its corresponding dimension. For example, when generating the question-criterion pairs for Idiom Interpretation, the LLM receives the prompt together with the idiom included, and its actual meaning to ensure it has the full background information to generate accurate questions and criteria. For Entity Reasoning, the explicit prompt with the specific entity name is provided. For Scientific Reasoning, the explicit prompt specifying the visible outcome derived from the scientific law is provided. As for Textual Image Design, rather than prescribing visual content, the core of evaluation is to assess whether generated images fulfill their intended communicative function. Therefore, only the original prompt is provided, this allows the T2I models to have full design freedom. 
The generated evaluation questions and criteria for the prompt ``Create a minimalist promotional poster for a workshop
on simplicity in design'' are shown in Table~\ref{tab:qc_2_example}, we can see they focus on evaluating functional elements including:
\begin{itemize}
    \item Clear presentation of workshop topic
    \item Inclusion of essential details (date/time/location)
    \item Consistency with minimalist design principles
\end{itemize}

\noindent \textbf{Templates for evaluation stage II: MLLM image evaluation.}
Table~\ref{tab:eval} presents the template used to evaluate the generated images for all four dimensions. Only the evaluation question-criterion pairs need to be replaced for each prompt. 

\subsection{Human Evaluation}
\label{human_evaluation}
The human evaluation is conducted on eight models: Stable-Diffusion-3-Medium, FLUX.1-schnell, HiDream-I1-full, Qwen-Image, Bagel, Janus-Pro-7B, GPT-Image-1, and Nano-Banana (4 diffusion models, 2 unified multimodal models, 2 proprietary model). 
We randomly select 20 prompts from each dimension and use these 8 models to generate 160 images per dimension. This results in 640 images in total for human evaluation. The evaluation is conducted by a group of volunteer postgraduate students. Three participants independently score each image, and we average their scores for each prompt-image pair.
The evaluation criteria provided to the human annotators are shown in Table~\ref{tab:c1}, ~\ref{tab:c2}, ~\ref{tab:c3}, and ~\ref{tab:c4}.
These criteria are tailored to each dimension in order to address different aspects. 
The correlations are calculated between the averaged human scores and the automatic metric score using Kendall's $\tau$ and Spearman's $\rho$.
The results are shown in Table~\ref{tab:human_corr_select}.

\begin{figure*}
\begin{tcolorbox}[
    width=\textwidth,
    colback=white,
    colframe=black,
    enhanced,
    sharp corners,
    boxrule=1pt,
    drop shadow,
    title=\textbf{Template for evaluation question-criterion generation: I. Idiom Interpretation},
    fonttitle=\bfseries\small,
    coltitle=white,
    valign=center
]

\texttt{<USER>:} 

I have a text-to-image generation model that can generate images based on given prompts. However, the model is not perfect and may fail to accurately capture the meaning of the prompt or depict it correctly.
Your task is to evaluate the generated image based on a specific prompt that contains an idiom.

Given the prompt: \{`id': \{\textcolor{cyan}{11}\}, `prompt': \{\textcolor{cyan}{After three weeks of burning the midnight oil, she finally submitted her dissertation.}\}, `idiom': \{\textcolor{cyan}{Burning the midnight oil}\}, 'idiom\_meaning': \{\textcolor{cyan}{To work or study late into the night, often requiring artificial light.}\}\}, 
you need to:

\begin{itemize}
    \item 1. identify what should be depicted in the image or the meaning the image should convey.
    \item 2. analyze the prompt and create a list of questions based on the key elements that the image should be checked against.
    \item 3. consider factors that could impact the aesthetics or visual quality of the image and list relevant questions.    
\end{itemize}

Please also design a scoring criterion for each question, where a score of 1 means ``yes (to the question)'', 0 means ``no'', and 0.5 means ``partially yes''.

Provide your answer in json format:

\begin{tcolorbox}[colback=gray!5, colframe=gray!30, boxrule=0.3pt, left=2mm, right=2mm, top=0.5mm, bottom=0.5mm, listing only]
\scriptsize
\begin{verbatim}
{
    'id': [prompt id], 
    'prompt': [the prompt], 
    'image_content': [what the image should convey], 
    'reason_evaluation': (
        here should be a dictionary with 3-5 pairs of question and criterion: 
        'q1': [question 1], 'c1': [criterion 1], 'q2': [question 2], 'c2': [criterion 2]...
    ), 
    'quality_evaluation': (same format as 'reason_evaluation' with 1-3 pairs of question and criterion)
}
\end{verbatim}
\end{tcolorbox}

\end{tcolorbox}

\caption{\textbf{Template used by the LLM to generate evaluation question-criterion pairs for the dimension of Idiom Interpretation.} The text highlighted in \textcolor{cyan}{cyan} should be replaced with details from the specific prompt.}
\label{tab:qc_1}
\end{figure*}

%%%%%%%%%%%%%%%%%%%%%%%%%%%%%%%%%%%%%%%%%%%%%
\begin{figure*}
\begin{tcolorbox}[
    width=\textwidth,
    colback=white,
    colframe=black,
    enhanced,
    sharp corners,
    boxrule=1pt,
    drop shadow,
    title=\textbf{Template for evaluation question-criterion generation: II. Textual Image Design},
    fonttitle=\bfseries\small,
    coltitle=white,
]

\texttt{<USER>:} 

I have a text-to-image generation model that can generate images based on given prompts. However, the model is not perfect and may fail to accurately reflect the prompt or depict the details correctly.

Given a prompt which is a design intention for a text-rich image like infographic or poster, your task is to evaluate whether the generated image correctly fulfill the design intention.

Here is the prompt: \{\{`id': \{\textcolor{cyan}{75}\}, `prompt': \{\textcolor{cyan}{Create a minimalist promotional poster for a workshop on simplicity in design.}\}\}, 
you need to:

\begin{itemize}
    \item 1. identify what should be depicted in the image and its functional purposes.
    \item 2. analyze the design intention and create a list of questions based on the key elements that the image should be checked against, including presence of required text elements.
    \item 3. consider factors that could impact the aesthetics or visual quality of the image and list relevant questions, such as overall layout, composition, color scheme, and consistency of visual style. Exclude all text-rendering and typographic aspects from these questions, as they should be evaluated separately.
    \item 4. consider factors that could impact the text rendering quality of the image and list relevant questions, such as text legibility (e.g., clear fonts and sufficient size), distortions or overlapping text, faulty text generation (e.g., misspellings and garbled characters), and overall typographic quality. Place all such aspects in \texttt{text\_rendering\_evaluation} instead of \texttt{quality\_evaluation}. Keep these questions distinct from whether the required information or text elements are present, and avoid evaluating the same aspect in more than one category.
\end{itemize}

Please also design a scoring criterion for each question, where a score of 1 means ``yes (to the question)'', 0 means ``no'', and 0.5 means ``partially yes''.

Provide your answer in json format:

\begin{tcolorbox}[colback=gray!5, colframe=gray!30, boxrule=0.3pt, left=2mm, right=2mm, top=0.5mm, bottom=0.5mm, listing only, listing options={breaklines=true}]
\scriptsize
\begin{verbatim}
{
    'id': [prompt id], 
    'prompt': [the prompt], 
    'image_content': [what the image should convey], 
    'reason_evaluation': (
        here should be a dictionary with 3-5 pairs of question and criterion: 
        'q1': [question 1], 'c1': [criterion 1], 'q2': [question 2], 'c2': [criterion 2]...
    ), 
    'quality_evaluation': (same format as 'reason_evaluation' with 1-3 question-criterion pairs),
    'text_rendering_evaluation': (same format as 'reason_evaluation' with 1-3 question-criterion pairs)
}
\end{verbatim}
\end{tcolorbox}

\end{tcolorbox}

\caption{\textbf{Template used by the LLM to generate evaluation question-criterion pairs for the dimension of Textual Image Design.} The text highlighted in \textcolor{cyan}{cyan} should be replaced with details from the specific prompt.}
\label{tab:qc_2}
\end{figure*}

%%%%%%%%%%%%%%%%%%%%%%%%%%%%%%%%%%%%%%%%%%%%%

\begin{figure*}
\begin{tcolorbox}[
    width=\textwidth,
    colback=white,
    colframe=black,
    enhanced,
    sharp corners,
    boxrule=1pt,
    drop shadow,
    title=\textbf{Template for evaluation question-criterion generation: III. Entity Reasoning},
    fonttitle=\bfseries\small,
    coltitle=white,
    valign=center
]

\texttt{<USER>:} 
I have a text-to-image generation model that can generate images based on given prompts. However, the prompts given to the model may contain implicit meanings or entities that are not directly stated.
Your task is to evaluate whether the generated image accurately represents the intended meaning of the prompt.

Given the prompt: \{\{`id': \{\textcolor{cyan}{31}\}, `prompt': \{\textcolor{cyan}{The CEO of Tesla unveiling the first electric car in 2008}\}, `explicit\_meaning': \{\textcolor{cyan}{Elon Musk unveiling the first electric car at Tesla in 2008}\}\}, 
you need to:

\begin{itemize}
\item 1. identify what should be depicted in the image in order to fully and accurately reflect the explicit meaning of the prompt.

\item 2. identify the entity that the model needs to infer from the prompt, and create a list of questions that check whether the image has correctly identified and depicted this entity.

\item 3. Consider other elements or details in the prompt (apart from the implicit entity), create a list of questions that check if the image accurately reflects these additional key elements.

\item 4. consider factors that could impact the aesthetics or visual quality of the image and list relevant questions.

\end{itemize}

Please also design a scoring criterion for each question, where a score of 1 means ``yes (to the question)'', 0 means ``no'', and 0.5 means ``partially yes''.

Provide your answer in json format:

\begin{tcolorbox}[colback=gray!5, colframe=gray!30, boxrule=0.3pt, left=2mm, right=2mm, top=0.5mm, bottom=0.5mm, listing only]
\scriptsize
\begin{verbatim}
{
    'id': [prompt id], 
    'prompt': [the prompt], 
    'explicit_meaning': [the explicit meaning], 
    'image_content': [what the image should convey], 
    'entity_evaluation': (
        here should be a dictionary with 1-3 pairs of question and criterion: 
        'q1': [question 1], 'c1': [criterion 1], 'q2': [question 2], 'c2': [criterion 2]...
    ), 
    'other_details_evaluation': (same format as 'entity_evaluation' with 1-3 pairs of question and criterion),
    'quality_evaluation': (same format as 'entity_evaluation' with 1-3 pairs of question and criterion)
}
\end{verbatim}
\end{tcolorbox}

\end{tcolorbox}

\caption{\textbf{Template used by the LLM to generate evaluation question-criterion pairs for the dimension of Entity Reasoning.} The text highlighted in \textcolor{cyan}{cyan} should be replaced with details from the specific prompt.}
\label{tab:qc_3}
\end{figure*}

%%%%%%%%%%%%%%%%%%%%%%%%%%%%%%%%%%%%%%%%%%%%%
\begin{figure*}
\begin{tcolorbox}[
    width=\textwidth,
    colback=white,
    colframe=black,
    enhanced,
    sharp corners,
    boxrule=1pt,
    drop shadow,
    title=\textbf{Template for evaluation question-criterion generation: IV. Scientific Reasoning},
    fonttitle=\bfseries\small,
    coltitle=white,
    valign=center
]

\texttt{<USER>:} 
I have a text-to-image generation model that can generate images based on given prompts. However, the prompts given to the model imply scientific laws (e.g., physics, chemistry, or biology) that can affect how the scene looks without explicit explanation.
Your task is to evaluate whether the generated image accurately reflects the scientific law and correctly portrays the resulting scene.

Given the prompt: \{\{`id': \{\textcolor{cyan}{34}\}, `prompt': \{\textcolor{cyan}{Two identical springs hang vertically from a support under the influence of Earth's gravity. One spring is unloaded, while the other supports the mass of a heavy ball. The entire system is shown at rest in a state of static equilibrium.}\}, `explicit\_meaning': \{\textcolor{cyan}{Two identical springs hang vertically from a support. The spring on the left is at its natural, un-stretched length. The spring on the right is stretched downwards by the weight of a heavy ball attached to its end. The entire system is motionless.}\}\},  
you need to:

\begin{itemize}

\item 1. describe what should be depicted in the image in order to fully and accurately reflect the explicit meaning of the prompt.

\item 2. identify any scientific law(s) that the model needs to infer from the prompt, and create a list of questions that check whether the image correctly demonstrates and complies with these scientific laws.

\item 3. consider other elements or details in the prompt that are not directly affected by the scientific law(s), create a list of questions that check if the image accurately represents these additional key elements.

\item 4. consider factors that could impact the aesthetics or visual quality of the image and list relevant questions.
\end{itemize}

Please also design a scoring criterion for each question, where a score of 1 means ``yes (to the question)'', 0 means ``no'', and 0.5 means ``partially yes''.

Provide your answer in json format:

\begin{tcolorbox}[colback=gray!5, colframe=gray!30, boxrule=0.3pt, left=2mm, right=2mm, top=0.5mm, bottom=0.5mm, listing only]
\scriptsize
\begin{verbatim}
{
    'id': [prompt id], 
    'prompt': [the prompt], 
    'explicit_meaning': [the explicit meaning], 
    'image_content': [what the image should depict], 
    'scientific_evaluation': (
        here should be a dictionary with 4-5 pairs of question and criterion: 
        'q1': [question 1], 'c1': [criterion 1], 'q2': [question 2], 'c2': [criterion 2]...
    ), 
    'other_details_evaluation': (same format as 'scientific_evaluation' with 4-5 question-criterion pairs),
    'quality_evaluation': (same format as 'scientific_evaluation' with 1-3 question-criterion pairs)
}
\end{verbatim}
\end{tcolorbox}

\end{tcolorbox}

\caption{\textbf{Template used by the LLM to generate evaluation question-criterion pairs for the dimension of Scientific Reasoning.} The text highlighted in \textcolor{cyan}{cyan} should be replaced with details from the specific prompt.}
\label{tab:qc_4}
\end{figure*}

%%%%%%%%%%%%%%%%%%%%%%%%%%%%%%%%%%%%%%%%%%%%%

\begin{figure*}
\begin{tcolorbox}[
    width=\textwidth,
    colback=white,
    colframe=black,
    enhanced,
    sharp corners,
    boxrule=1pt,
    drop shadow,
    title=\textbf{Example of generated evaluation questions and criteria: II. Textual Image Design},
    fonttitle=\bfseries\small,
    coltitle=white,
    valign=center
]

\small
\raggedright
\texttt{<ASSISTANT>:}\par
\{\par
\texttt{``id''}: 75,\par
\texttt{``prompt''}: ``Create a minimalist promotional poster for a workshop on simplicity in design.'',\par
\texttt{``image\_content''}: ``A clean, minimalist promotional poster visually communicating a workshop focused on simplicity in design. ...'',\par
\smallskip\texttt{``reason\_evaluation''}: \{\par
\texttt{``q1''}: ``Does the poster clearly communicate that it is promoting a workshop, not just a generic design or quote?'',\par
\texttt{``c1''}: \{ ``1'': ``The image explicitly presents itself as a workshop promotion (e.g., includes the word 'workshop' or equivalent and event-related framing).'', ``0.5'': ``The image implies an event or learning session but is ambiguous about being a workshop.'', ``0'': ``The image does not suggest any kind of event or workshop context.'' \},\par
\texttt{``q2''}: ``Is the theme of 'simplicity in design' clearly conveyed through the visual concept and messaging?'',\par
\texttt{``c2''}: \{ ``1'': ``The image explicitly references simplicity in design (through text such as 'simplicity', 'minimalism', or similar) and visually embodies minimalism.'', ``0.5'': ``Either the text or visuals suggest simplicity/minimalism, but the connection to 'simplicity in design' as a workshop theme is weak or indirect.'', ``0'': ``The image does not clearly relate to simplicity or minimalism as a design theme.'' \},\par
\texttt{``q3''}: ``Are essential promotional details for a workshop present (e.g., title, date/time, and some indication of location or format)?'',\par
\texttt{``c3''}: \{ ``1'': ``The poster includes at least a clear workshop title plus date/time and location or format (e.g., online).'', ``0.5'': ``Some key details are present (e.g., title and date) but at least one essential detail (date/time or location/format) is missing or vague.'', ``0'': ``The poster lacks basic event details, making it unclear how or when the workshop occurs.'' \},\par
\ldots\par
\},\par
\smallskip\texttt{``text\_rendering\_evaluation''}: \{\par
\texttt{``q1''}: ``Is all visible text legible, with characters rendered clearly and at a sufficient size to be read without strain?'',\par
\texttt{``c1''}: \{ ``1'': ``All text is clearly rendered and easily readable at normal viewing size.'', ``0.5'': ``Most text is legible, but some smaller or faint text is harder to read.'', ``0'': ``Significant portions of text are illegible due to size, blurring, or poor rendering.'' \},\par
\texttt{``q2''}: ``Is the text free from noticeable distortions, overlapping, or artifacts (e.g., warped letters, text merging with graphics)?'',\par
\texttt{``c2''}: \{ ``1'': ``Text appears clean, with no significant distortions, overlaps, or visual artifacts.'', ``0.5'': ``Minor distortions or overlaps exist but do not seriously hinder readability.'', ``0'': ``Text is heavily distorted, overlapped, or merged with other elements, impairing readability.'' \},\par
\texttt{``q3''}: ``Is the generated text largely free of misspellings, random characters, or garbled words that would confuse the viewer?'',\par
\texttt{``c3''}: \{ ``1'': ``Text is grammatically coherent and free of obvious misspellings or random/garbled sequences.'', ``0.5'': ``There are occasional minor spelling errors or odd characters, but the main information remains understandable.'', ``0'': ``Text contains frequent misspellings, random strings, or garbled characters that make the content difficult to interpret.'' \},\par
\},\par
\smallskip\texttt{``quality\_evaluation''}: \{\par
\texttt{``q1''}: ``Does the overall layout and composition support a minimalist aesthetic (e.g., balanced use of negative space, few visual elements, clear hierarchy)?'',\par
\texttt{``c1''}: \{ ``1'': ``The composition is clean and balanced, with ample negative space and a small number of well-organized visual elements.'', ``0.5'': ``The composition shows some minimalist intent but has minor clutter or imbalance that weakens the aesthetic.'', ``0'': ``The composition is busy, cluttered, or poorly balanced, conflicting with minimalism.'' \},\par
\ldots\par
\},\par
\}\par
\end{tcolorbox}

\caption{Excerpt of the LLM-generated question-criterion pairs for Textual Image Design (prompt 75), supplied to the MLLM for image evaluation.}
\label{tab:qc_2_example}
\end{figure*}

\begin{figure*}
\begin{tcolorbox}[
    width=\textwidth,
    colback=white,
    colframe=black,
    enhanced,
    sharp corners,
    boxrule=1pt,
    drop shadow,
    title=\textbf{Template for image evaluation},
    fonttitle=\bfseries\small,
    coltitle=white,
    valign=center
]

\texttt{<USER>:} Describe this image.

\texttt{<ASSISTANT>:} 
The image depicts a group of four people in a meeting setting, with one person standing and the others seated around a table. The standing individual, dressed in a dark suit and light blue shirt, is playfully pointing at the nose of one of the seated individuals...

\texttt{<USER>:} 
Based on the image and your previous description, answer the following questions: q1, q2, ...
For each question, assign a score of 1, 0.5 or 0 according to the corresponding scoring criteria: c1, c2, ...

Here are the questions and criteria: \{\textcolor{cyan}{accuracy\_evaluation\_qc}\}.

Carefully consider the image and each question before responding, then provide your answer in json format:
\{\{``reason'': [your detailed reasoning], ``score'': [s1,s2, ...]\}\}

\texttt{<ASSISTANT>:} 
\{
  ``reason'': [
    ``q1: The image clearly shows a meeting environment with a conference table, office-like setting, and individuals in professional attire. This aligns well with the criteria for a clear meeting setting.'',
    
    ``q2: One person is visibly acting as the joke-teller by making a playful gesture towards another person's face, which suggests an active role in engaging the group. This meets the criteria for a clear speaker with an active role.'',
    
    ``q3: The audience members are displaying relaxed and post-icebreaking reactions, such as genuine smiles and laughter, indicating that the tension has been relieved. This fulfills the criteria for clear reactions showing tension relief.'',
    
    ``q4: The idiom 'break the ice' is depicted entirely metaphorically through the playful interaction and the text overlay, without any literal elements like actual ice. This satisfies the criteria for an entirely metaphorical depiction.''
  ],
  ``score'': [1, 1, 1, 1]
\}

\texttt{<USER>:} 
Based on the image and your previous description, answer the following questions: q1, q2, ...
For each question, assign a score of 1, 0.5 or 0 according to the corresponding scoring criteria: c1, c2, ...

Here are the questions and criteria: \{\textcolor{cyan}{quality\_evaluation\_qc}\}

Carefully consider the image and each question before responding, then provide your answer in json format:
\{\{``reason'': [your detailed reasoning], ``score'': [s1,s2, ...]\}\}

\texttt{<ASSISTANT>:} 
\{
  ``reason'': [
    ``The image is visually clear with well-defined subjects. There is no noticeable blur or distortion, allowing for easy identification of the individuals and their actions. Therefore, it scores high on clarity.'',
    
    ``The composition is balanced, with the speaker positioned centrally and the audience arranged cohesively around the table. There is no awkward cropping, and all elements contribute to a harmonious arrangement.'',
    
    ``The colors and lighting are appropriate, enhancing the mood of the scene. Warm tones and even lighting create a positive and inviting atmosphere, supporting the lighthearted interaction depicted.''
  ],
  ``score'': [1, 1, 1]
\}

\end{tcolorbox}

\caption{\textbf{Template used by MLLM for image evaluation.} The text highlighted in \textcolor{cyan}{cyan} should be replaced with the specific evaluation question and criterion pairs for the given prompt. An example output for Idiom Interpretation is also provided. }
\label{tab:eval}
\end{figure*}

\begin{figure*}
\begin{tcolorbox}[
    width=\textwidth,
    colback=white,
    colframe=black,
    enhanced,
    sharp corners,
    boxrule=1pt,
    drop shadow,
    title=\textbf{Human Evaluation Criteria: I. Idiom Interpretation},
    fonttitle=\bfseries\small,
    coltitle=white,
    valign=center
]

Given the prompt ``\{\textcolor{cyan}{prompt}\}'', idiom it contains ``\{\textcolor{cyan}{idiom}\}'' and idiom meaning ``\{\textcolor{cyan}{idiom meaning}\}'', please rate the alignment between the image and the prompt on a scale of 1 to 5 according to the criteria:

\begin{itemize}
\item 5 - The image independently and unambiguously depicts all elements of the prompt (actions, emotions, context, and consequences) without requiring additional text for further explanation. Annotators can instantly recognize the intended message and its nuances.

\item 4 - The image clearly reflects the core idea of the prompt, capturing major elements (e.g., key actions, settings) but may lack subtle details (e.g., context, specific emotions). Annotators can easily connect it to the prompt with minimal effort.

\item 3 - The image partially represents the prompt, focusing on generic aspects (e.g., basic scenario) but missing critical details (e.g., cause-effect relationships, tone, implied consequences). Annotators can only understand the link after reading the prompt and idiom meaning.

\item 2 - The image vaguely or superficially relates to the prompt, with weak or unclear ties to its specifics (e.g., missing context, conflicting tone, wrong elements). Even with the prompt, the connection feels unclear or underdeveloped.

\item 1 - The image contradicts or ignores the prompt’s core message (e.g., misrepresenting outcomes, tone, or relationships). Annotators can find it irrelevant or misleading, even with the prompt. 
\end{itemize}

\end{tcolorbox}

\caption{\textbf{Human Evaluation Criterion for Idiom Interpretation.} The text highlighted in \textcolor{cyan}{cyan} should be replaced with information about the specific prompt.}
\label{tab:c1}
\end{figure*}

\begin{figure*}
\begin{tcolorbox}[
    width=\textwidth,
    colback=white,
    colframe=black,
    enhanced,
    sharp corners,
    boxrule=1pt,
    drop shadow,
    title=\textbf{Human Evaluation Criteria: II. Textual Image Design},
    fonttitle=\bfseries\small,
    coltitle=white,
    valign=center
]
Given a prompt describing a design intention for a rich-text image ``\{\textcolor{cyan}{prompt}\}'', please rate how well the image reflects the design prompt on a scale of 1 to 5 according to the criteria:

\begin{itemize}
    \item 5 – Exemplary Alignment: The image perfectly reflects the design prompt, addressing all specified elements (e.g., text type, visuals, data, tone), delivers the core message clearly, and has no flaws (no errors, coherent emphasis, and non-superficial intentions fully realized).
    
    \item 4 – Good Alignment with Minor Gaps: The image aligns well with the prompt, fulfills core requirements, and conveys the message effectively but has minor oversights (e.g., missing details, slight color/text inconsistency) that do not undermine the overall intent.
    
    \item 3 – Partial Fulfillment: The image captures the general idea and addresses key aspects of the prompt (e.g., correct type, basic message) but overlooks or misrepresents notable details (e.g., incorrect text/data visualization, inconsistent tone) or contains errors affecting clarity.

    \item 2 – Superficial Compliance: The image only superficially resembles the prompt’s intent (e.g., correct theme but missing critical elements like key visuals, misaligned focus, or unaddressed design implications) and may include distracting errors or inconsistencies.

    \item 1 – Mismatched or Incomplete: The image fails to address the prompt’s requirements (e.g., wrong image type, missing core message, major design inaccuracies) with pervasive errors, rendering it ineffective or off-topic.

\end{itemize}

\end{tcolorbox}
\caption{\textbf{Human Evaluation Criterion for  Textual Image Design.} The text highlighted in \textcolor{cyan}{cyan} should be replaced with information about the specific prompt.}
\label{tab:c2}
\end{figure*}

\begin{figure*}
\begin{tcolorbox}[
    width=\textwidth,
    colback=white,
    colframe=black,
    enhanced,
    sharp corners,
    boxrule=1pt,
    drop shadow,
    title=\textbf{Human Evaluation Criteria: III. Entity Reasoning},
    fonttitle=\bfseries\small,
    coltitle=white,
    valign=center
]

Given the prompt ``\{\textcolor{cyan}{prompt}\}'' and the actual entity it indicates ``\{\textcolor{cyan}{explicit\_prompts}\}'', please rate the alignment between the image and the prompt on a scale of 1 to 5 according to the criteria:
\begin{itemize}
\item 5 - Perfectly alignment: the image faithfully captures all key elements of the prompt (subject, setting, time period, distinguishing features) with no inaccuracies.

\item 4 - Mostly accurate: the image depicts core elements correctly but has minor errors (e.g., slight anachronisms, missing details, or incomplete context).

\item 3 - Partially correct: the image includes some relevant elements but mixes in inaccuracies (e.g., wrong context, missing critical details, or moderate deviations from the prompt).

\item 2 - Weak representation: the image only loosely connected to the prompt, with significant inaccuracies (e.g., wrong subject identity, era or location).

\item 1 - Completely inaccurate: the image fails to reflect the prompt’s core theme, details, or context (e.g., unrelated subject, fantasy elements, or contradictory visuals).

\end{itemize}

Please carefully examine the image and check if all the details in the prompt are correctly addressed in the image.

\end{tcolorbox}
\caption{\textbf{Human Evaluation Criterion for Entity Reasoning.} The text highlighted in \textcolor{cyan}{cyan} should be replaced with information about the specific prompt.}
\label{tab:c3}
\end{figure*}

\begin{figure*}
\begin{tcolorbox}[
    width=\textwidth,
    colback=white,
    colframe=black,
    enhanced,
    sharp corners,
    boxrule=1pt,
    drop shadow,
    title=\textbf{Human Evaluation Criteria: IV. Scientific Reasoning},
    fonttitle=\bfseries\small,
    coltitle=white,
    valign=center
]

Given a prompt that relates to scientific laws ``\{\textcolor{cyan}{prompt}\}'' and its reference prompt ``\{\textcolor{cyan}{explicit\_prompt}\}'', please rate the image on a scale of 1 to 5 according to the criteria:

\begin{itemize}
\item 5 - Excellent: The image accurately depicts all the elements from the prompt (subject, action, setting, state) and strictly adheres to scientific laws. No errors in details or logic.

\item 4 - Good: The image includes all key elements from the prompt but has minor scientific inaccuracies or small missing details.

\item 3 - Fair: The image includes most elements but has moderate errors: either missing an critical element or clearly violating scientific principles.

\item 2 - Poor: The image omits multiple key elements and has significant scientific inaccuracies.

\item 1 - Fail: The image fails to represent the prompt (e.g., incorrect subjects/actions) and completely ignores scientific laws.

\end{itemize}

Please carefully examine the image and check if the image correctly address the scientific law inherent in the prompt.

\end{tcolorbox}
\caption{\textbf{Human Evaluation Criterion for Scientific Reasoning.} The text highlighted in \textcolor{cyan}{cyan} should be replaced with information about the specific prompt.}
\label{tab:c4}
\end{figure*}

\end{document}